\documentclass[lettersize,journal]{IEEEtran}
\usepackage[T1]{fontenc}
\usepackage{amsmath,amsfonts}
\usepackage{algorithmic}
\usepackage{algorithm}
\usepackage{array}
\usepackage[caption=false,font=normalsize,labelfont=sf,textfont=sf]{subfig}
\usepackage{textcomp}
\usepackage{stfloats}
\usepackage{url}
\usepackage{verbatim}
\usepackage{graphicx}
\usepackage{cite}

\usepackage{amsmath,amsfonts,array,textcomp,stfloats,url,graphicx,cite,booktabs,multirow}
\usepackage[caption=false,font=normalsize,labelfont=sf,textfont=sf]{subfig}
\begin{document}

\title{Single-step Retrosynthesis via \\ Reinforcement Learning from Forward Feedback}

\author{Zixun Lan$^{1}$, Zhenfu Liu$^{2}$, Yi Li$^{1}$, Xinheng Wang$^{1}$, Mingmin Chi$^{3}$, Qibiao Wang$^{2}$, Jie Zhu$^{4}$, Xiaohui Tian$^{4}$, \\Fei Ma$^{1*}$\\

\thanks{$^*$corresponding author}

\thanks{$^1$ Xi’an Jiaotong-Liverpool University, Suzhou 215000, China}

\thanks{$^2$ Sichuan University of Science \& Engineering, Zigong 643000, China}
\thanks{$^3$ Fudan University, Shanghai 200433, China}
\thanks{$^4$ Sinochem Digital Intelligence Technology Co., Ltd., Beijing, China}

\thanks{This paper was produced by the IEEE Publication Technology Group. They are in Piscataway, NJ.}
\thanks{Manuscript received April 19, 2021; revised August 16, 2021.}}

\markboth{Journal of \LaTeX\ Class Files,~Vol.~14, No.~8, August~2021}%
{Shell \MakeLowercase{\textit{et al.}}: A Sample Article Using IEEEtran.cls for IEEE Journals}


\maketitle

\begin{abstract}
Single-step retrosynthesis is inherently a one-to-many inverse problem, yet most models are trained and evaluated against a single recorded reactant set. Reference-based training provides no explicit positive signal for unrecorded alternatives, while exact-match evaluation counts non-matching predictions as incorrect regardless of their chemical plausibility. We propose Retrosynthesis via Reinforcement Learning from Forward Feedback (RetroRLFF), which optimizes reactant generation using forward consistency with the target product rather than reference agreement alone. Starting from a supervised decoder-only Transformer, RetroRLFF generates complete reactant candidates and uses a frozen forward reaction model to predict products for comparison with the target. A sequence-level reward retains reference agreement to preserve recovery of recorded reactions and adds forward consistency to provide feedback for non-reference candidates, while penalizing invalid outputs. Direct Advantage-Based Policy Optimization then updates the retrosynthesis policy from this feedback without requiring human preference annotations or a preference-trained reward model. On USPTO-50K, RetroRLFF outperforms all 14 baselines in candidate-level round-trip accuracy at $k\in\{1,3,5,10\}$. With approximately 50 million parameters, it reaches 90.1\% top-1 round-trip accuracy, exceeding R-SMILES, the strongest competing baseline on this metric, by 8.5 percentage points. Relative to its supervised policy, round-trip accuracy improves from 83.1\% to 90.1\%, while exact-match accuracy is maintained at 57.9\% versus 57.5\%. To complement forward-model-based evaluation, we conduct an anonymized expert assessment on 131 randomly selected test cases, in which organic synthesis experts prefer RetroRLFF over EditRetro in 67.9\% of comparisons. These results support forward feedback as an effective training signal for improving forward consistency while retaining top-1 reference recovery. Expert assessment provides complementary evidence of expert-assessed synthetic feasibility on the selected cases, although forward consistency and expert preference do not establish experimental feasibility.
\end{abstract}


\section{Introduction}

Single-step retrosynthesis prediction aims to identify possible reactant sets that can generate a target product. In computer-aided synthesis planning (CASP), these predictions define the precursor branches explored during multistep search, thereby affecting search efficiency and the quality of the proposed routes \cite{F02,T04}. CASP supports applications in drug discovery, materials design, and automated chemical experiments.

Existing research has mainly developed three technical approaches to this task. Template-based methods extract local transformation rules from historical reactions and then select applicable templates through similarity retrieval, template classification, or neural symbolic reasoning \cite{T01,T02,T03,T04,T05,T06}. Semi-template-based methods typically first identify the reaction center, decompose the target product into synthons---hypothetical precursor fragments obtained by disconnecting the product---and then complete these fragments into reactants \cite{ST01,ST02,ST03,lan2025retrosynthesis,lan2024rcsearcher,zhu2023single}. Template-free methods do not rely on explicit reaction templates but directly learn the mapping between molecular sequences or molecular graphs. These template-free methods have expanded from SMILES-based autoregressive generation and graph editing to pre-training, diffusion models, and discrete flow matching \cite{TF01,TF02,TF03,TF04,TF05,TF06,N09,N04,N05,TF11}.

\begin{figure}[!t]
    \centering
    \includegraphics[width=\columnwidth]
        {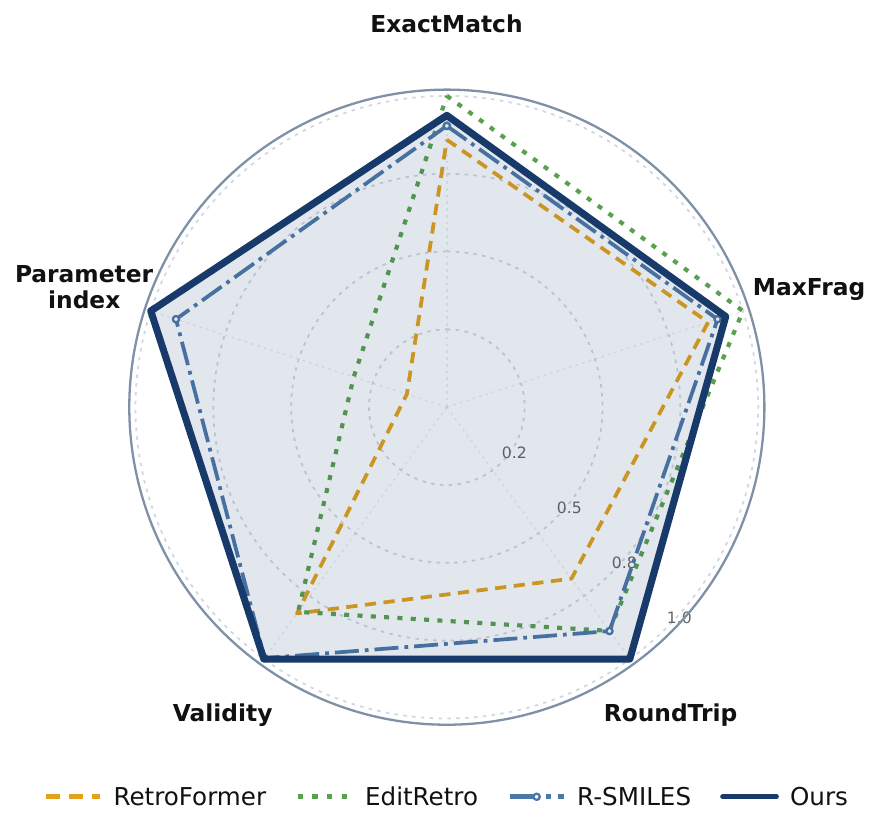}
    \caption{Normalized comparison of RetroRLFF (Ours) with three state-of-the-art retrosynthesis models on USPTO-50K across four prediction metrics and a parameter index. The parameter index is proportional to the reciprocal of the retrosynthesis model's parameter count; higher values indicate smaller models. Higher scores are better on all axes. RetroRLFF achieves higher round-trip consistency with fewer parameters.}
    \label{fig:model-metrics-radar}
\end{figure}

Despite these architectural differences, most of the methods discussed above follow a similar supervision paradigm. Template classification loss, sequence cross-entropy, and graph-edit likelihood differ in form, but their supervision is centered on the reference reactions recorded in the dataset. Model performance is also commonly evaluated using exact-match accuracy, which measures whether the predicted reactant set matches the recorded reference. Thus, both training and evaluation primarily emphasize recovery of the recorded solutions.

However, single-step retrosynthesis is inherently a one-to-many inverse problem: a target product can often be synthesized from several different reactant sets, whereas patent datasets usually record only one or a few of these options. A prediction that does not match the recorded reference is therefore not necessarily chemically invalid. Manual evaluation and error analyses have shown that some non-matching predictions still represent plausible reaction schemes \cite{T01,ST04,ST05,TF03}. Reference-based training provides no explicit positive signal for these unrecorded alternatives, while exact-match evaluation counts them as incorrect. Consequently, reference agreement does not fully capture the practical goal of identifying reactants that can produce the target product. This motivates a complementary training signal that can assess candidate reactants against the target product without requiring them to match the recorded reference.

To address the mismatch between training and application objectives, we propose Retrosynthesis via Reinforcement Learning from Forward Feedback (RetroRLFF). The framework uses forward reaction prediction to assess consistency between candidate reactants and the target product. Compared with inferring reactants from a product, predicting the major product from a given set of reactants generally involves less ambiguity \cite{TF01}, and models such as the Molecular Transformer have demonstrated strong forward prediction performance \cite{F01}. These observations motivate using forward prediction as scalable proxy feedback, although agreement between the predicted product and the target does not establish experimental feasibility. Round-trip consistency has been used to evaluate retrosynthetic candidates \cite{F02,T04,ST04,ST06,TF06,TF11}; RetroRLFF incorporates it into policy optimization for complete reactant-set generation.

Starting from a supervised decoder-only policy, RetroRLFF samples complete reactant candidates for each target product. A frozen forward reaction model predicts a product from each candidate for comparison with the target. We retain reference agreement to preserve recovery of recorded reactions, while forward consistency provides an additional learning signal for candidates that differ from the recorded reference. The sequence-level reward combines these two signals using canonicalized comparisons, with negative penalties for invalid reactants, invalid forward products, and processing failures. It requires neither human preference annotations nor a preference-trained reward model. Direct Advantage-Based Policy Optimization (DAPO) translates this feedback into token-level policy updates through critic-estimated advantages, with the supervised policy serving as a fixed reference \cite{DAPO}.

\begin{figure*}[!t]
    \centering
    \includegraphics[width=\textwidth]
        {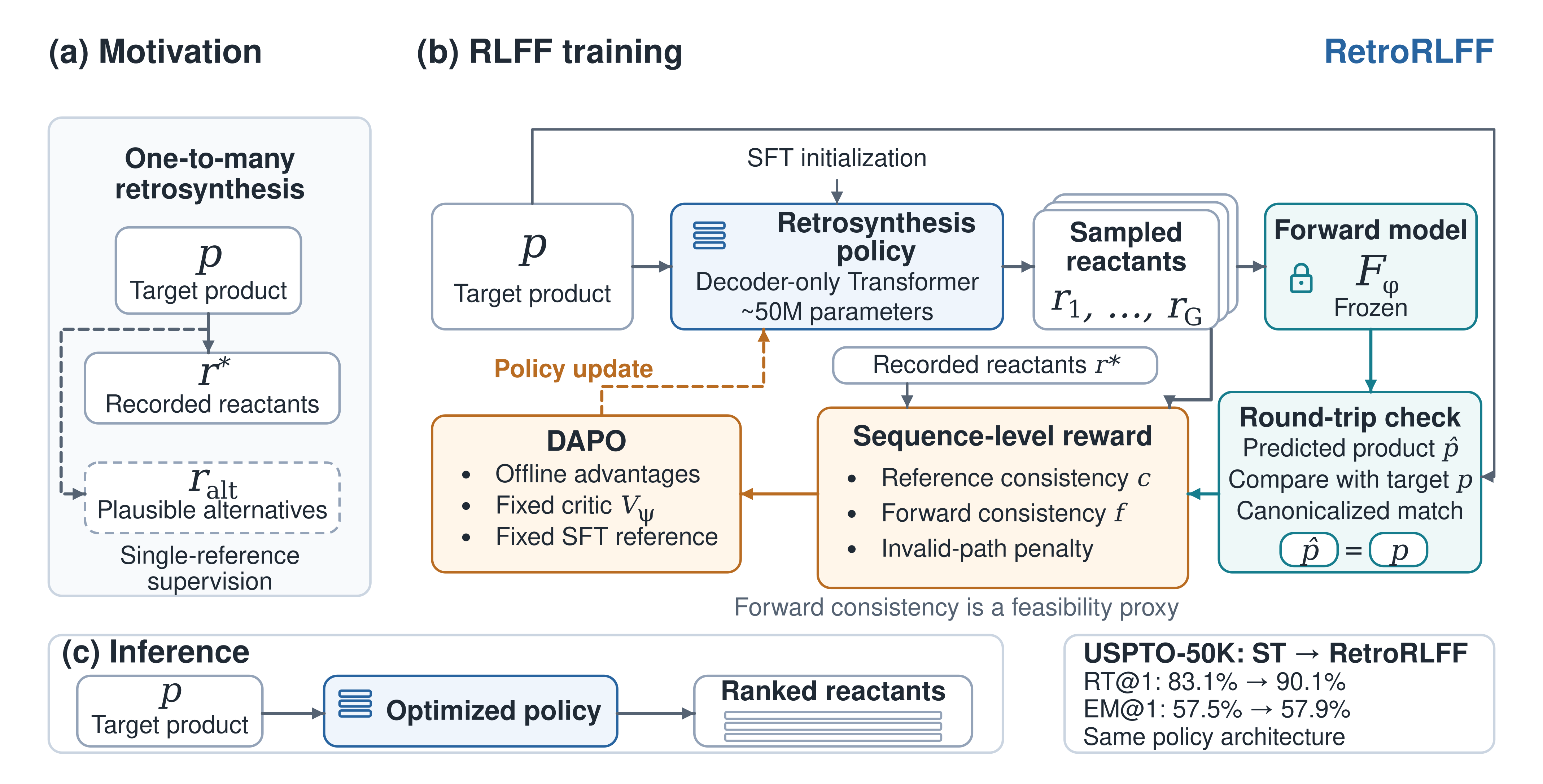}
    \caption{Overview of RetroRLFF.
    (a) Single-step retrosynthesis is a one-to-many inverse problem:
    plausible reactant alternatives may differ from the recorded reference.
    (b) A supervised decoder-only policy samples complete reactant candidates,
    which are evaluated by a frozen forward model.
    Canonicalized reference agreement and forward consistency determine
    a sequence-level reward, with penalties for invalid evaluation paths.
    Direct Advantage-Based Policy Optimization (DAPO) updates the policy
    using offline advantages while keeping the critic and supervised
    reference policy fixed.
    (c) The optimized policy generates ranked reactant candidates.
    On USPTO-50K, RLFF improves top-1 round-trip accuracy from
    83.1\% to 90.1\% while retaining exact-match accuracy
    (57.5\% versus 57.9\%) with the same policy architecture.
    Forward consistency is a proxy for synthetic feasibility,
    rather than experimental validation.}
    \label{fig:retrorlff-overview}
\end{figure*}

The same frozen forward predictor supplies the consistency component of the training reward and computes round-trip accuracy during evaluation. We assess subsequent RLFF optimization by comparing the supervised and optimized policies while keeping their architecture and parameter count unchanged. Both use a standard decoder-only Transformer with approximately 50 million parameters and no reaction templates, graph-editing modules, or explicit reasoning components. As summarized in Fig.~\ref{fig:model-metrics-radar}, RetroRLFF also achieves higher round-trip consistency than the three compared baselines while using a smaller retrosynthesis model. Fig.~\ref{fig:retrorlff-overview} summarizes the motivation, training process, and inference procedure.

We report candidate-level round-trip accuracy at $k\in\{1,3,5,10\}$, defined as the fraction of all top-$k$ reactant candidates whose forward-predicted products match their respective targets after canonicalization. On USPTO-50K, RetroRLFF achieves the highest round-trip accuracy among the compared methods at every evaluated cutoff. Relative to the supervised policy, RLFF improves top-1 round-trip accuracy from 83.1\% to 90.1\%, while top-1 exact-match accuracy changes from 57.5\% to 57.9\%. Reward-configuration analysis further reveals tradeoffs between reference recovery and forward consistency.

Because round-trip consistency remains a model-based proxy for synthetic feasibility, we complement it with anonymized expert rankings based on synthetic feasibility. On a randomly selected subset of 131 USPTO-50K test cases, organic synthesis experts prefer RetroRLFF over EditRetro in 67.9\% of comparisons. Qualitative case studies further illustrate plausible non-reference reactant alternatives. Together, these evaluations support improved forward consistency and the plausibility of selected predictions, while neither round-trip consistency nor expert preference establishes experimental success.

In summary, we outline our primary contributions as follows:
\begin{itemize}

\item We propose RetroRLFF, a reinforcement learning framework that incorporates forward reaction feedback into single-step retrosynthesis. It combines reference agreement with forward consistency to update the retrosynthesis policy, providing an additional learning signal for non-reference candidates without requiring human preference annotations.

\item We demonstrate the effectiveness of RetroRLFF using a standard decoder-only Transformer with approximately 50M parameters. On USPTO-50K, it achieves the highest round-trip accuracy among the compared methods at all reported cutoffs; relative to its supervised policy, top-1 round-trip accuracy increases from 83.1\% to 90.1\%, while exact-match accuracy changes from 57.5\% to 57.9\%.

\item We complement benchmark evaluation with anonymized expert assessment, qualitative analyses of non-reference reactant alternatives, and illustrations of recursive precursor expansion. On the randomly selected subset of 131 test cases, experts prefer RetroRLFF over EditRetro in 67.9\% of comparisons and over the recorded reference in 71.8\%, providing complementary evidence beyond exact reference matching.
\end{itemize}

\section{Related Work}
\subsection{Template-Based and Semi-Template-Based Methods}

Template-based methods represent a reaction as a local subgraph transformation. RetroSim retrieves reaction templates from similar reactions in the training set. NeuralSym formulates template selection as a neural classification problem, while GLN jointly models template applicability and the corresponding reactants \cite{T01,T02,T03}. LocalRetro replaces reaction-level templates with local atom and bond transformations \cite{T04}. MHNreact learns associations between products and templates with modern Hopfield networks, whereas RetroKNN uses local nearest-neighbor retrieval to improve predictions for less frequent reactions \cite{T05,T06}. Reaction templates provide explicit transformation rules and reduce the search space. However, their performance depends strongly on the quality and coverage of template extraction. Long-tailed template frequencies, costly subgraph matching, and the choice of template granularity can also limit generalization to unseen transformations.

Semi-template-based methods preserve the chemically meaningful process of bond disconnection followed by reactant completion. G2Gs and RetroXpert first identify reaction centers and split the product into synthons. They then reconstruct the reactants through graph generation or Transformer-based completion \cite{ST01,ST02}. GraphRetro uses stepwise graph transformations, Graph2Edits generates an autoregressive sequence of graph edits, and Retro-MTGR introduces multitask graph representation learning \cite{ST03,ST06,ST09}. These structural priors make intermediate decisions easier to interpret. However, an incorrect reaction center can propagate errors to the completion stage. Dependence on atom mapping and predefined edit sets also constrains the possible outputs. Moreover, each stage is usually supervised with the recorded bond disconnection or reactant set. The decomposition therefore does not remove the bias caused by single-reference supervision.

\subsection{Template-Free Retrosynthesis Methods}

Sequence-based methods formulate retrosynthesis as a machine translation problem. The product SMILES is treated as the source sequence, and the reactant SMILES form the target sequence. Early seq2seq models used recurrent neural networks, whereas SCROP combined a Transformer with an auxiliary SMILES correction module \cite{TF01,TF02}. SMILES augmentation exposes the model to multiple valid representations of the same molecule and reduces its dependence on one fixed serialization \cite{TF10}. R-SMILES aligns the root atoms on both sides of a reaction, thereby shortening the distance between corresponding atoms \cite{TF05}. RetroFormer introduces local--global attention to capture both reaction-center information and global molecular context \cite{TF06}. Chemformer first learns general chemical representations from unlabeled SMILES and is then fine-tuned for reaction prediction and retrosynthesis \cite{N09}. These methods are flexible and easy to scale. However, autoregressive decoding remains sensitive to token order, suffers from exposure bias, and may propagate early errors through the remaining sequence.

Graph-based methods operate more directly on molecular topology. MEGAN represents a reaction as a sequence of graph-edit operations \cite{TF03}. Graph2SMILES combines a permutation-invariant graph encoder with a sequence decoder \cite{TF04}. G2GT instead uses graph-to-graph attention and self-training to generate reactant graphs \cite{TF08}. These models preserve structural information more explicitly and reduce their dependence on arbitrary SMILES order. Their main difficulties arise from the discrete action space, the ordering of graph edits, and the cost of searching over possible transformations.

Recent work has also explored non-autoregressive and iterative generative processes. EditRetro generates reactants by repeatedly applying explicit Levenshtein editing operations to the product SMILES \cite{TF09}. Although it formulates this process as a Markov decision process, its reward is defined by the distance to the recorded reactant sequence. It therefore remains tied to reference supervision rather than forward reaction feedback. RetroBridge constructs a Markov bridge between product and reactant graph distributions \cite{N04}. DiffER uses categorical diffusion \cite{N05}, while RetroSynFlow adopts discrete flow matching \cite{TF11}. These approaches broaden the available sampling mechanisms and provide more control over candidate diversity. Nevertheless, their basic training objectives still rely mainly on paired product--reactant records. RetroSynFlow additionally uses Feynman--Kac steering with sequential Monte Carlo resampling and a learned forward-synthesis reward oracle. This signal is applied only during inference and does not update the parameters of the trained generator.

\subsection{Reinforcement Learning from Human and Model Feedback}

Reinforcement learning makes it possible to optimize a reward defined over a complete output rather than the likelihood of individual reference tokens. Early preference-based reinforcement learning learned reward functions from pairwise human comparisons and then optimized a policy against the learned reward \cite{RL07}. This idea was later applied to language generation. Stiennon et al.\ trained a summarization model with a reward learned from human preferences \cite{RL08}. InstructGPT adopted a widely used three-stage pipeline consisting of supervised fine-tuning, reward-model training from ranked outputs, and policy optimization with reinforcement learning \cite{RL02}. Proximal Policy Optimization (PPO) is commonly used in the final stage because its clipped objective limits the size of each policy update \cite{RL01}. A Kullback--Leibler divergence penalty can further keep the optimized policy close to the supervised reference model \cite{RL02}. In this way, RLHF moves the training signal from reproducing one reference output toward evaluating complete generations according to human preferences.

Human preference labels are costly and difficult to collect at scale. Reinforcement learning from AI feedback (RLAIF) addresses this limitation by replacing some human rankings with preferences produced by another model \cite{RL06}. A standard RLAIF pipeline uses these preferences to train a reward model. Direct RLAIF instead obtains rewards directly from an AI evaluator during policy optimization, without fitting a separate reward model \cite{RL06}. This idea can be extended to task-specific evaluators that provide automatic feedback. However, the evaluator remains an imperfect proxy for the true objective. Strong optimization may exploit its errors, causing the proxy reward to increase while actual output quality stops improving or even decreases \cite{RL09}.

Model-based feedback has also been introduced into chemistry and retrosynthesis. RLSynC uses a standalone forward synthesis model to reward synthon completion within an offline--online multi-agent reinforcement learning framework \cite{RL03}. RetroGFN trains a GFlowNet with a learned feasibility proxy so that it can sample diverse reactions beyond those recorded in the training set \cite{N03}. More broadly, ether0 applies reinforcement learning across a large collection of experimentally grounded chemistry tasks \cite{RL04}. Round-Trip Reinforcement Learning uses the success of a forward--reverse transformation as a reward for improving the self-consistency of chemical language models \cite{RL05}. These studies show that chemical models can learn from automatic evaluators rather than from exact reference matching alone. However, they differ in the generated object, the source of the reward, and the policy optimization algorithm.

RetroRLFF follows this feedback-driven view but is not RLHF in the strict sense. It does not require human preference comparisons or a reward model trained to reproduce such preferences. Given a target product $p_i$, the retrosynthesis policy generates a complete candidate reactant string $\hat r_{i,j}$. A forward reaction model then maps $\hat r_{i,j}$ to a predicted product $\hat p_{i,j}=F_\phi(\hat r_{i,j})$. Agreement between $\hat p_{i,j}$ and $p_i$ provides the forward-consistency component of the reward used for DAPO. Unlike post-hoc round-trip evaluation or the inference-time steering used by RetroSynFlow, this reward directly updates the parameters of the retrosynthesis policy \cite{TF11}. Unlike RLSynC, RetroRLFF optimizes complete reactant generation rather than only synthon completion. It is therefore more accurately described as reinforcement learning from forward-model feedback. The resulting reward measures forward-model-based consistency, not experimentally verified reaction feasibility.






\section{Methodology}

Fig.~\ref{fig:retrorlff-overview} summarizes RetroRLFF.
We first formulate single-step retrosynthesis and define
reference-based and forward-consistency metrics.
We then initialize a decoder-only policy through supervised
training and optimize it using forward-feedback rewards
and Direct Advantage-Based Policy Optimization.
The following subsections detail the task, supervised
initialization, reward construction, and policy update.

\subsection{Task Definition}

Let $\mathcal{D}=\{(p_i,r_i^\ast)\}_{i=1}^{N}$ denote a dataset of recorded reactions. Here, $p_i$ is the SMILES of the target product, $r_i^\ast$ is the dot-separated SMILES of its reference reactants, and $N$ is the number of test samples. 

A retrosynthesis model with parameters $\theta$ defines a conditional policy $\pi_\theta(r\mid p)$. Given $p_i$, it produces an ordered list of $K$ candidates,
\begin{equation}
\hat{\mathcal{R}}_i^{K}
=
(\hat r_{i,1},\hat r_{i,2},\ldots,\hat r_{i,K}),
\end{equation}
where $\hat r_{i,j}$ is the $j$-th candidate ranked by the model score.

We use $\mathcal{C}(\cdot)$ to denote SMILES canonicalization. For a string containing several dot-separated molecules, $\mathcal{C}(\cdot)$ canonicalizes each molecule separately, sorts the resulting component strings, and then joins them with dots. The comparison is therefore independent of molecule order. We further define $\mathcal{V}(s)\in\{0,1\}$ as an indicator of whether every molecule in $s$ can be parsed and canonicalized. Single-step retrosynthesis predicts a set of possible reactants from one product. Since several reactant sets may produce the same product, $r_i^\ast$ is a recorded reference rather than the only possible chemical solution.

\subsection{Evaluation Metric}

We report both reference-based exact matching accuracy and forward-model-based round-trip accuracy. Round-trip results can depend on candidate aggregation, forward decoding, and the treatment of invalid outputs. These differences can prevent direct comparison across studies. We therefore fix the complete evaluation protocol described below.

\subsubsection{Exact Matching Accuracy}

Each test product is assigned $K$ ranked reactant candidates. For $1\leq k\leq K$, Top-$k$ exact matching accuracy is defined as
\begin{equation}
\operatorname{EM@}k
=
\frac{1}{N}
\sum_{i=1}^{N}
\mathbb{I}
\left[
\exists j\leq k:
\mathcal{V}(\hat r_{i,j})=1
\land
\mathcal{C}(\hat r_{i,j})
=
\mathcal{C}(r_i^\ast)
\right],
\end{equation}
where $\mathbb{I}[\cdot]$ is the indicator function. A test sample is correct if at least one of its first $k$ candidates exactly matches the reference after canonicalization.

\subsubsection{Round-Trip Accuracy}

Let $F_\phi$ be a forward reaction prediction model with fixed parameters $\phi$. It maps each predicted reactant string to one product string:
\begin{equation}
\hat p_{i,j}=F_\phi(\hat r_{i,j}).
\end{equation}
The forward model always uses greedy decoding with \texttt{beam\_size = 1} and \texttt{do\_sample = False}. Thus, each reactant candidate produces exactly one deterministic forward prediction.

We define RT@$k$ as a candidate-level metric:
\begin{equation}
\begin{aligned}
\mathrm{RT@}k
&=
\frac{1}{Nk}
\sum_{i=1}^{N}
\sum_{j=1}^{k}
\mathbb{I}\Bigl[
\mathcal{V}(\hat r_{i,j})
\land
\mathcal{V}(\hat p_{i,j})
\\[-2pt]
&\qquad\qquad
\land
\mathcal{C}(\hat p_{i,j})
=
\mathcal{C}(p_i)
\Bigr].
\end{aligned}
\end{equation}
Each reactant candidate is evaluated independently. It is a round-trip success if its forward prediction matches the original target product after canonicalization. The denominator is $Nk$, so all top-$k$ candidates from all samples receive equal weight. Invalid reactants, invalid forward products, parsing errors, and failed model calls are counted as failures. This detailed protocol is intended to reduce differences caused by inconsistent implementations of round-trip accuracy and to support fair comparisons across studies.

\subsection{Pretraining}

Let $\mathcal{D}_{\mathrm{tr}}=\{(p_i,r_i^\ast)\}_{i=1}^{M}$ be the training set containing $M$ reaction pairs. We use a standard decoder-only Transformer to learn the mapping: $p_i \longrightarrow r_i^\ast$.

Both sides are represented by their SMILES strings. Let $\mathcal{T}(\cdot)$ denote the tokenizer. The tokenized product and reference reactants are
\begin{equation}
x_i=\mathcal{T}(p_i),
\qquad
y_i=\mathcal{T}(r_i^\ast)
=
(y_{i,1},\ldots,y_{i,T_i}),
\end{equation}
where $T_i$ is the number of reactant tokens. Conditioned on $x_i$ and the previously generated tokens $y_{i,<t}$, the model predicts the next reactant token. The supervision training objective is
\begin{equation}
\mathcal{L}_{\mathrm{ST}}(\theta)
=
-\frac{1}{M}
\sum_{i=1}^{M}
\sum_{t=1}^{T_i}
\log
\pi_\theta
\left(
y_{i,t}
\mid
x_i,y_{i,<t}
\right).
\end{equation}
The optimized parameters are denoted by $\theta_{\mathrm{ST}}$.

We deliberately use a simple decoder-only architecture. It contains no reaction templates, graph modules, or additional handcrafted features. Its parameter count is kept below that of the SOTA single-step retrosynthesis models. This controlled setup reduces the effects of model scale and specialized components. It allows us to focus on how RLFF changes reference consistency and synthetic feasibility. The SMILES augmentation and reranking strategy follows common settings from previous work \cite{TF10,TF05}.

\subsection{Reinforcement Learning from Forward Feedback}

RLFF starts from the pretrained supervision policy. For each product, the retrosynthesis policy samples a group of complete reactant candidates. The fixed forward model then evaluates each candidate against the target product. Sampling is used for retrosynthesis rollouts, whereas $F_\phi$ always uses deterministic greedy decoding.

\subsubsection{Reward Function}

The reward combines reference consistency and forward consistency. For a candidate reactant string $r$, reference consistency is
\begin{equation}
c_i(r)
=
\mathbb{I}
\left[
\mathcal{C}(r)
=
\mathcal{C}(r_i^\ast)
\right].
\end{equation}
Forward consistency is defined as
\begin{equation}
f_i(r)
=
\mathbb{I}
\left[
\mathcal{C}(F_\phi(r))
=
\mathcal{C}(p_i)
\right].
\end{equation}
It serves as a proxy for synthetic feasibility under the forward model, but it does not prove that the reaction will succeed experimentally.

We also define the validity of the complete evaluation path:
\begin{equation}
v_i(r)
=
\mathbb{I}
\left[
\mathcal{V}(r)=1
\land
\mathcal{V}(F_\phi(r))=1
\right].
\end{equation}
A failed forward call is treated as an invalid output, and any canonicalized equality involving an invalid string is false. The reward is
\begin{equation}
R_i(r)
=
\begin{cases}
r_{\mathrm{inv}},
&
v_i(r)=0,
\\[4pt]
\alpha c_i(r)
+
\beta f_i(r)
+
\gamma c_i(r)f_i(r)
+
\delta,
&
v_i(r)=1,
\end{cases}
\end{equation}
where $r_{\mathrm{inv}}<0$ penalizes invalid outputs and processing errors. The nonnegative coefficients $\alpha$ and $\beta$ weight reference agreement and forward consistency, respectively; $\gamma$ controls their interaction, and $\delta$ is a constant offset for valid evaluation paths.

A valid candidate receives $\alpha+\beta+\gamma+\delta$ when both conditions hold, $\alpha+\delta$ for reference-only agreement, $\beta+\delta$ for forward-only agreement, and $\delta$ when neither condition holds.

\subsubsection{Direct Advantage-Based Policy Optimization}

The sequence-level reward defined above is used as the terminal signal for policy optimization. Following DAPO \cite{DAPO}, we first form an offline advantage dataset from sampled rollouts. A state $s$ contains the target product and the current reactant prefix, while an action $a$ is the next reactant token. A separately trained critic $V_\psi$ estimates the value of the resulting prefix from completions scored by $R_i(r)$. The advantage $\widehat{A}(s,a)$ is obtained by centering these values over the candidate actions sampled for the same state. This provides step-specific credit instead of assigning the same terminal reward to every token.

We set the frozen reference policy to $\pi_{\mathrm{ref}}=\pi_{\theta_{\mathrm{ST}}}$ and minimize
\begin{equation}
\begin{aligned}
\mathcal{L}_{\mathrm{DAPO}}(\theta)
&=
\frac{1}{2}
\mathbb{E}_{(s,a,\widehat{A})\sim\mathcal{D}_{\mathrm{adv}}}
\\[-2pt]
&\quad
\left[
\left(
\frac{\widehat{A}(s,a)}{\tau}
-
\log
\frac{\pi_\theta(a\mid s)}
{\pi_{\mathrm{ref}}(a\mid s)}
\right)^2
\right].
\end{aligned}
\end{equation}
Here, $\mathcal{D}_{\mathrm{adv}}$ contains offline state--action--advantage tuples rather than pairwise preference labels, and $\tau>0$ is the temperature coefficient controlling deviation from the reference policy. The loss fits the log probability ratio to the scaled advantage, increasing the probability of advantageous actions and suppressing disadvantageous ones. We update $\theta$ by gradient descent while keeping $\pi_{\mathrm{ref}}$ and $V_\psi$ fixed.

\section{Experiments}
\label{sec:experiments}

We evaluate RetroRLFF from seven complementary perspectives:

(1) We compare overall performance on USPTO-50K using Exact Matching Accuracy, Round-Trip Accuracy, and Maximal Fragment Accuracy at $k\in\{1,3,5,10\}$. These metrics assess reference recovery and candidate-level forward consistency, providing complementary views of prediction quality.

(2) We compare the supervised training policy with three subsequent RLFF configurations to assess the effect of reinforcement learning and different reward coefficients. The analysis examines the balance between reference agreement and forward consistency while keeping the interaction coefficient, constant offset, and invalid-output penalty fixed.

(3) We evaluate round-trip accuracy across non-ring, ring-opening, and ring-forming reactions, comparing RetroRLFF with R-SMILES, EditRetro, G2Retro, and RetroFormer to examine performance across different structural transformations and candidate ranks.

(4) We investigate architectural sensitivity by varying the parameter count, number of Transformer layers, and number of attention heads, assessing the stability of both exact-match and round-trip performance across five configurations.

(5) We conduct an anonymized human expert evaluation on 131 randomly selected test instances. Experts rank the RetroRLFF predictions, EditRetro predictions, and recorded reference reactants by synthetic feasibility, allowing ties. We report rank distributions and pairwise preferences.

(6) We recursively apply RetroRLFF to six representative target molecules to illustrate successive precursor predictions, candidate ranks, and branching patterns in multistep retrosynthetic exploration.

(7) We examine five qualitative cases in which RetroRLFF proposes alternatives to the recorded reactants, discussing their chemical plausibility, potential synthetic advantages, and limitations beyond exact reference matching.

\subsection{Dataset}
\label{subsec:dataset}

We use USPTO-50K, a standard benchmark for single-step retrosynthesis derived from reactions in United States patents \cite{TF01,T01,T03}. The curated set contains 50,016 atom-mapped reactions assigned to ten broad classes: heteroatom alkylation and arylation, acylation and related processes, carbon--carbon bond formation, heterocycle formation, protections, deprotections, reductions, oxidations, functional group interconversion, and functional group addition.

Following the fixed protocol used by the Conditional Graph Logic Network (GLN) \cite{T03}, we use 40,008 reactions for training, 5,001 for validation, and 5,007 for testing. These subsets account for approximately 80\%, 10\%, and 10\% of the benchmark, respectively. For evaluation, dot-separated reactants are processed using the component-wise canonicalization $\mathcal{C}(\cdot)$ defined in the Methodology section.


\subsection{Baselines}
We compare RetroRLFF with the 14 retrosynthesis baselines reported in Table~\ref{tab:benchmark}, covering template-based, semi-template-based, and template-free approaches. Template-based baselines include RetroSim \cite{T01}, NeuralSym \cite{T02}, GLN \cite{T03}, and LocalRetro \cite{T04}. The comparison also includes RetroXpert \cite{ST02}, GraphRetro \cite{ST03}, RetroPrime \cite{ST04}, and G2Retro \cite{ST05}, together with MEGAN \cite{TF03}, GTA \cite{GTA}, RetroFormer \cite{TF06}, Graph2Edits \cite{ST06}, R-SMILES \cite{TF05}, and EditRetro \cite{TF09}, spanning reaction-center prediction, graph editing, and sequence generation. For reaction-group analysis, we compare against R-SMILES, EditRetro, G2Retro, and RetroFormer. EditRetro serves as the competing model in human expert evaluation. We additionally use our supervision training policy (ST, without RL) as the internal baseline for assessing subsequent RLFF optimization.

\subsection{Evaluation Metrics}
\label{subsec:evaluation-metrics}

We use the notation established in the Methodology section. Top-$k$ Exact Matching Accuracy ($\mathrm{EM@}k$) measures reference consistency: a test product is correct if at least one of its first $k$ ranked candidates is valid and equals the recorded reactant set after canonicalization. Top-$k$ Round-Trip Accuracy ($\mathrm{RT@}k$) measures forward-model-based synthetic feasibility: every one of the $Nk$ candidates is evaluated independently, and a success requires the deterministic forward prediction to equal the original product. Their complete definitions, invalid-output rules, and normalization conventions are given in the preceding evaluation-metric subsection and are not repeated here. The forward model $F_\phi$ is a pretrained Molecular Transformer \cite{F01}. The same frozen checkpoint is used for both training-reward computation and round-trip evaluation.

In this paper, the definitions of the other two evaluation metrics are as follows:

\subsubsection{Maximal Fragment Accuracy}
\label{subsubsec:maxfrag}

Maximal Fragment Accuracy (MaxFrag) evaluates whether a prediction recovers the principal reactant fragment, while ignoring differences in the remaining components \cite{TF10}. Let $\mathcal{M}(s)$ select the component of a valid dot-separated reactant string $s$ whose canonical SMILES has the greatest character length. Top-$k$ MaxFrag Accuracy is
\begin{equation}
\begin{aligned}
\mathrm{MaxFrag@}k
&=
\frac{1}{N}
\sum_{i=1}^{N}
\mathbb{I}\Bigl[
\exists j\leq k:
\mathcal{V}(\hat r_{i,j})=1
\\[-2pt]
&\qquad\qquad\land
\mathcal{C}\bigl(\mathcal{M}(\hat r_{i,j})\bigr)
=
\mathcal{C}\bigl(\mathcal{M}(r_i^\ast)\bigr)
\Bigr].
\end{aligned}
\label{eq:maxfrag-at-k}
\end{equation}
The symbols $N$, $\hat r_{i,j}$, $r_i^\ast$, $\mathcal{V}$, $\mathcal{C}$, and $\mathbb{I}$ have the same meanings as in the Methodology section. Invalid predictions cannot produce a match. A larger $\mathrm{MaxFrag@}k$ is better; it indicates that the main recorded reactant is recovered by at least one candidate within the first $k$ ranks.

\subsubsection{Validity}
\label{subsubsec:validity}

Validity measures whether the model generates reactant SMILES that can be processed as molecular structures. Following the definition of $\mathcal{V}(\cdot)$ in the Methodology section, Top-$k$ Validity is defined as
\begin{equation}
\mathrm{Valid@}k
=
\frac{1}{Nk}
\sum_{i=1}^{N}
\sum_{j=1}^{k}
\mathbb{I}
\left[
\mathcal{V}(\hat r_{i,j})=1
\right].
\label{eq:validity-at-k}
\end{equation}
A candidate is valid if RDKit can parse and canonicalize every molecule in its dot-separated reactant string. Each of the first $k$ candidates is checked independently. The denominator is $Nk$, so every candidate has the same weight, and an invalid candidate is counted as a failure. A higher $\mathrm{Valid@}k$ means that the model produces fewer malformed reactant SMILES.

\begin{table*}[!t]
\caption{Overall performance on the USPTO-50K test set. RoundTrip reports candidate-level forward-model consistency among the first $k$ predictions ($\mathrm{RT@}k$); ExactMatch and MaxFrag report the proportions of products with at least one matching prediction among the first $k$ ($\mathrm{EM@}k$ and $\mathrm{MaxFrag@}k$). All entries are percentages. Bold values indicate the highest result in each column.}
\label{tab:benchmark}
\centering
\begingroup
\scriptsize
\renewcommand{\arraystretch}{1.04}
\setlength{\tabcolsep}{1.8pt}
\begin{tabular*}{\textwidth}{@{\extracolsep{\fill}}lcccccccccccc@{}}
\toprule
\multirow{3}{*}{\textbf{Model}}
& \multicolumn{12}{c}{\textbf{Top-$k$ Performance (\%)}} \\
\cmidrule(lr){2-13}
& \multicolumn{4}{c}{\textbf{RoundTrip}}
& \multicolumn{4}{c}{\textbf{ExactMatch}}
& \multicolumn{4}{c}{\textbf{MaxFrag}} \\
\cmidrule(lr){2-5}\cmidrule(lr){6-9}\cmidrule(lr){10-13}
& \textbf{1} & \textbf{3} & \textbf{5} & \textbf{10}
& \textbf{1} & \textbf{3} & \textbf{5} & \textbf{10}
& \textbf{1} & \textbf{3} & \textbf{5} & \textbf{10} \\
\midrule
\multicolumn{13}{l}{\textbf{Template-Based}} \\
RetroSim        & 61.1 & 55.3 & 52.4 & 47.6 & 37.3 & 54.7 & 63.3 & 74.1 & 40.5 & 58.9 & 67.9 & 78.7 \\
NeuralSym       & 66.2 & 67.2 & 64.8 & 51.4 & 44.4 & 65.3 & 72.4 & 78.9 & 53.6 & 74.6 & 81.5 & 87.8 \\
GLN             & 66.0 & 63.2 & 61.4 & 57.4 & 52.5 & 74.6 & 80.5 & 86.9 & 57.2 & 72.4 & 78.2 & 84.0 \\
LocalRetro      & 65.6 & 61.9 & 59.8 & 56.0 & 53.1 & 76.1 & 84.4 & \textbf{91.0} & 57.4 & 79.4 & 86.9 & \textbf{92.8} \\
\midrule
\multicolumn{13}{l}{\textbf{Semi-Template-Based}} \\
RetroXpert      & 66.5 & 42.8 & 36.1 & 30.3 & 50.4 & 61.1 & 62.3 & 63.4 & 59.1 & 73.7 & 78.0 & 83.1 \\
GraphRetro      & 65.1 & 59.5 & 55.5 & 48.5 & 53.7 & 68.3 & 72.2 & 75.5 & 58.5 & 71.6 & 75.0 & 77.7 \\
RetroPrime      & 80.4 & 62.6 & 53.9 & 43.3 & 51.4 & 70.8 & 74.0 & 76.1 & 56.7 & 75.3 & 79.5 & 81.4 \\
G2Retro         & 63.7 & 58.1 & 54.1 & 46.3 & 54.1 & 74.1 & 81.2 & 86.7 & 56.4 & 75.7 & 81.4 & 86.5 \\
\midrule
\multicolumn{13}{l}{\textbf{Template-Free}} \\
MEGAN            & 65.6 & 61.0 & 57.6 & 51.0 & 48.1 & 70.7 & 78.4 & 86.1 & 53.3 & 74.9 & 82.3 & 88.5 \\
GTA              & 65.5 & 54.2 & 47.2 & 36.8 & 51.1 & 67.6 & 74.8 & 81.6 & 56.7 & 75.4 & 79.9 & 81.7 \\
RetroFormer      & 64.8 & 60.6 & 57.2 & 49.1 & 53.2 & 71.1 & 76.6 & 82.1 & 58.0 & 70.8 & 74.8 & 78.4 \\
Graph2Edits      & 64.9 & 58.8 & 55.3 & 48.9 & 54.9 & 76.7 & 83.0 & 89.0 & 59.2 & 79.7 & 85.7 & 90.6 \\
R-SMILES         & 81.6 & 74.5 & 69.9 & 59.1 & 55.7 & 77.1 & 84.1 & 90.4 & 60.1 & 80.2 & 86.7 & 92.2 \\
EditRetro        & 81.5 & 67.7 & 58.6 & 44.4 & \textbf{60.8} & \textbf{80.6} & \textbf{86.0} & 90.3 & \textbf{65.8} & \textbf{84.7} & \textbf{89.0} & 92.3 \\
\midrule
\textbf{Ours}   & \textbf{90.1} & \textbf{78.3} & \textbf{72.0} & \textbf{62.7} & 57.9 & 74.5 & 82.8 & 86.4 & 61.2 & 78.2 & 84.7 & 88.3 \\
\bottomrule
\end{tabular*}
\endgroup
\end{table*}

\subsection{Implementation Details} 
RetroRLFF uses a standard decoder-only Transformer with six layers, eight attention heads, and approximately 50 million parameters. The architecture is retained during supervision training (ST) and subsequent RLFF optimization, so the gains over the supervised policy are obtained without increasing its parameter count. Unless otherwise specified, we set \(\alpha=1.8\) and \(\beta=1.2\) for reference consistency and forward consistency, respectively. The interaction coefficient \(\gamma\) and constant offset \(\delta\) are both set to \(-1\); invalid evaluation paths receive the direct penalty \(r_{\mathrm{inv}}=-1\). The frozen forward model uses deterministic greedy decoding.

ST runs on two Intel Xeon Gold 6133 CPUs at 2.50 GHz and four NVIDIA GeForce RTX 4090 GPUs, each with 24 GB of memory. RLFF runs on eight Intel Xeon Gold 6240 CPUs at 2.60 GHz and twelve NVIDIA Quadro RTX 6000 GPUs with 24 GB each; nine GPUs support policy fine-tuning and sampling, while three perform forward-model scoring. Early stopping occurs at 325,000 ST steps and 130,000 RLFF steps.

\subsection{Results and Discussion}
\label{subsec:results-and-discussion}

\subsubsection{Overall Performance Comparison}
\label{subsubsec:overall-performance}

Table~\ref{tab:benchmark} compares RetroRLFF with template-based, semi-template-based, and template-free methods on the USPTO-50K test set described in Section~\ref{subsec:dataset}. We report $k\in\{1,3,5,10\}$ for three complementary criteria. The forward model evaluates each reactant candidate separately for $\mathrm{RT@}k$, whereas $\mathrm{EM@}k$ and $\mathrm{MaxFrag@}k$ count a test product once if any of its first $k$ predictions matches the recorded full reactant set or its maximal fragment, respectively (Section~\ref{subsec:evaluation-metrics}). Consequently, the decreasing $\mathrm{RT@}k$ values at larger $k$ reflect the average consistency of a wider candidate pool; they are not a top-$k$ success rate over products.

RetroRLFF obtains the highest round-trip consistency at every reported cutoff: 90.1\%, 78.3\%, 72.0\%, and 62.7\% at $k=1,3,5,10$, respectively. The strongest competing round-trip results are 81.6\%, 74.5\%, 69.9\%, and 59.1\% from R-SMILES, so the corresponding absolute gains are 8.5, 3.8, 2.1, and 3.6 percentage points. In particular, the 90.1\% top-1 result indicates that the first-ranked reactant set more frequently regenerates the target under the fixed forward model. The advantage persists when lower-ranked candidates are included, although the average round-trip rate decreases for RetroRLFF from 90.1\% at $k=1$ to 62.7\% at $k=10$.

The reference-based metrics give a more qualified picture. RetroRLFF achieves 57.9\% $\mathrm{EM@}1$ and 61.2\% $\mathrm{MaxFrag@}1$, compared with EditRetro's leading 60.8\% and 65.8\%, respectively. At $k=10$, RetroRLFF reaches 86.4\% exact match and 88.3\% maximal-fragment accuracy, while LocalRetro reaches 91.0\% and 92.8\%. Thus, the round-trip gains do not imply a gain in recovering the single recorded reactant set or its largest fragment. The distinct patterns are consistent with the motivation for forward feedback: a prediction can agree with the target under the forward model without reproducing the recorded reference.

\begin{figure*}[!t]
\centering
\includegraphics[width=0.92\textwidth]{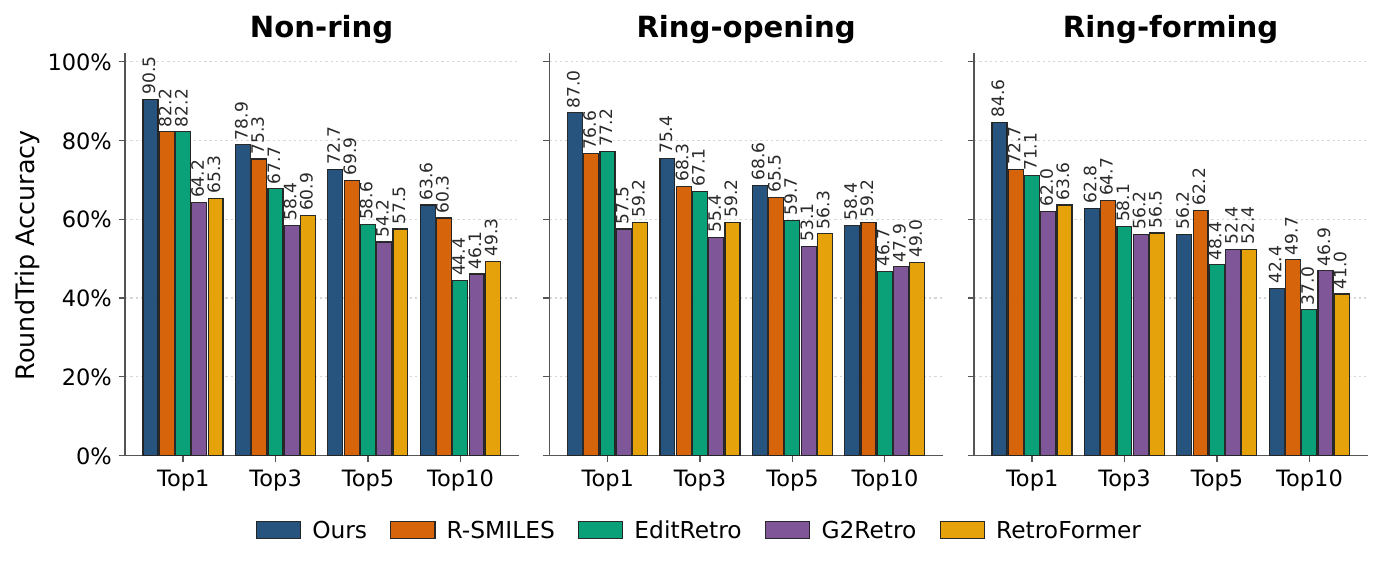}
\caption{Candidate-level round-trip accuracy ($\mathrm{RT@}k$, \%) on USPTO-50K for non-ring, ring-opening, and ring-forming reactions at $k\in\{1,3,5,10\}$.}
\label{fig:class_accuracy}
\end{figure*}

\subsubsection{Ablation Studies and Reward Coefficient Analysis}
\label{subsubsec:ablation-reward}


Table~\ref{tab:sft_rl_top1} compares supervised training (ST, without RL) with three RLFF configurations on USPTO-50K, using top-1 Exact Matching Accuracy ($\mathrm{EM@}1$) and Round-Trip Accuracy ($\mathrm{RT@}1$). ST+RL denotes subsequent optimization of the supervised policy through Direct Advantage-Based Policy Optimization. All three ST+RL configurations fix $\gamma=\delta=r_{\mathrm{inv}}=-1$ and vary only $\alpha$ and $\beta$; invalid evaluation paths therefore receive the same penalty. This comparison examines the RL stage and coefficient settings rather than isolating individual reward terms.

\begin{table}[!t]
\caption{Top-1 exact-match and round-trip accuracy on USPTO-50K for the supervised policy and three RLFF reward configurations. All ST+RL configurations use $\gamma=\delta=r_{\mathrm{inv}}=-1$. Coefficient entries are not applicable to ST.}
\label{tab:sft_rl_top1}
\centering
\begingroup
\scriptsize
\renewcommand{\arraystretch}{1.2}
\setlength{\tabcolsep}{3pt}
\begin{tabular*}{\columnwidth}{@{\extracolsep{\fill}}lcccc@{}}
\toprule
\textbf{Configuration} & $\boldsymbol{\alpha}$ & $\boldsymbol{\beta}$ & \textbf{ExactMatch} & \textbf{RoundTrip} \\
\midrule
ST (w/o RL) & --- & --- & 57.5\% & 83.1\% \\
ST+RL & 1.8 & 1.2 & 57.9\% & 90.1\% \\
ST+RL & 2 & 1 & 48.7\% & 93.8\% \\
ST+RL & 2 & 2 & 52.4\% & 88.3\% \\
\bottomrule
\end{tabular*}
\endgroup
\end{table}

With $(\alpha,\beta)=(1.8,1.2)$, RLFF improves round-trip accuracy from 83.1\% to 90.1\%, a gain of 7.0 percentage points, while exact-match accuracy changes from 57.5\% to 57.9\%. This setting therefore improves forward consistency while retaining the observed level of reference recovery, matching the reported RetroRLFF results in Table~\ref{tab:benchmark}.

For valid candidates, the reward vectors for $(c_i(r),f_i(r))=(1,1),(1,0),(0,1),(0,0)$ are $(1,0.8,0.2,-1)$, $(1,1,0,-1)$, and $(2,1,1,-1)$ under $(\alpha,\beta)=(1.8,1.2),(2,1),(2,2)$, respectively. The first setting rewards forward-consistent non-reference predictions while favoring reference agreement. The second assigns zero reward to forward-only agreement; the third rewards either condition equally and increases the reward for joint agreement.

The $(\alpha,\beta)=(2,1)$ setting achieves the highest round-trip accuracy, 93.8\%, but reduces exact match to 48.7\%. Increasing $\beta$ to 2 while keeping $\alpha=2$ raises exact match to 52.4\% while lowering round-trip accuracy to 88.3\%. Thus, a larger feasibility coefficient does not necessarily improve the resulting round-trip performance. These results indicate sensitivity to the full reward configuration, without establishing the underlying optimization mechanism. Among the tested settings, $(\alpha,\beta)=(1.8,1.2)$ achieves the highest exact-match accuracy and outperforms $(\alpha,\beta)=(2,2)$ on both metrics, supporting its use when preserving reference recovery is also required.

\begin{figure*}[!t]
\centering
\includegraphics[width=0.90\textwidth]{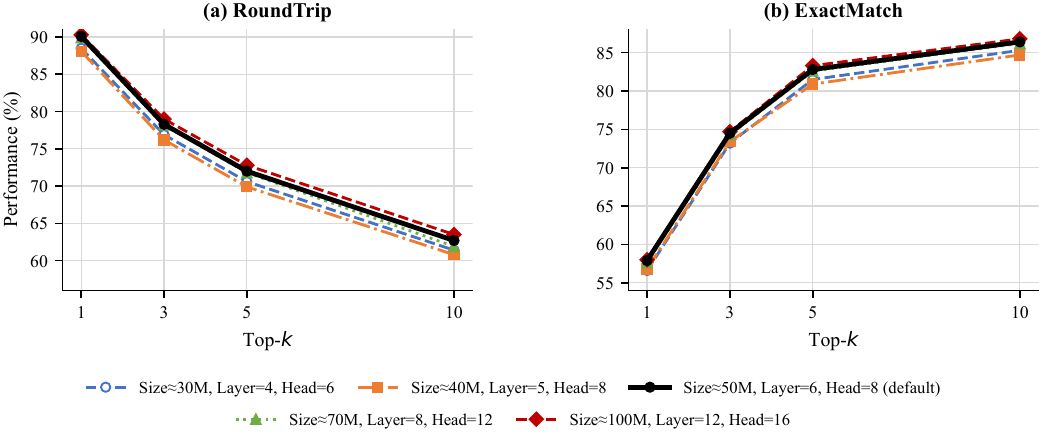}
\caption{Sensitivity of RetroRLFF to model size, layer count, and attention head count on USPTO-50K: (a) candidate-level round-trip accuracy ($\mathrm{RT@}k$) and (b) product-level exact matching accuracy ($\mathrm{EM@}k$), for $k\in\{1,3,5,10\}$. The legend specifies each architecture; the default configuration has approximately 50M parameters, six layers, and eight attention heads.}
\label{fig:Robustness}
\end{figure*}

\subsubsection{Performance across Reaction Types}
\label{subsubsec:reaction-types}

Fig.~\ref{fig:class_accuracy} compares RetroRLFF with R-SMILES, EditRetro, G2Retro, and RetroFormer across three reaction groups. We use the candidate-level $\mathrm{RT@}k$ defined in Section~\ref{subsec:evaluation-metrics}, evaluated within each group.

RetroRLFF achieves the highest $\mathrm{RT@}1$ for non-ring, ring-opening, and ring-forming reactions, reaching 90.5\%, 87.0\%, and 84.6\%, respectively. These results exceed the strongest baseline in each group by 8.3, 9.8, and 11.9 percentage points. It also leads at $k=3$, with 78.9\%, 75.4\%, and 64.7\%. The advantage for leading predictions therefore extends across all three groups, including both types of ring transformations.

At larger cutoffs, the advantage depends on the reaction group. RetroRLFF remains best for non-ring reactions at $k=5$ and $k=10$ (72.7\% and 63.6\%). For ring-opening reactions, it leads at $k=5$ but slightly trails R-SMILES at $k=10$ (58.4\% versus 59.2\%). For ring-forming reactions, its 56.2\% and 42.4\% at $k=5$ and $k=10$ are below R-SMILES's 62.2\% and 49.7\%. This identifies forward consistency among lower-ranked ring-forming candidates as a remaining limitation. Since $\mathrm{RT@}k$ averages candidate outcomes, its decline reflects reduced average consistency as the pool expands.

\subsubsection{Hyperparameter Sensitivity and Robustness}
\label{subsubsec:hyperparameter-robustness}

We examine architectural sensitivity by comparing five RetroRLFF configurations in Fig.~\ref{fig:Robustness}. Their approximate parameter counts, layer counts, and attention head counts are $(30\mathrm{M},4,6)$, $(40\mathrm{M},5,8)$, $(50\mathrm{M},6,8)$, $(70\mathrm{M},8,12)$, and $(100\mathrm{M},12,16)$, respectively. The 50M configuration is the default. We assess both forward consistency and reference recovery using the metrics defined in Section~\ref{subsec:evaluation-metrics}.

The curves remain close across all evaluated cutoffs. Approximate readings from the figure place $\mathrm{RT@}1$ at 88--90\% and $\mathrm{RT@}10$ at 61--64\%, while $\mathrm{EM@}1$ is about 57--58\% and $\mathrm{EM@}10$ about 85--87\%. The 100M model performs best on both metrics, but its advantage over the default configuration is less than approximately one percentage point at each cutoff. Increasing capacity therefore provides only modest additional accuracy within the tested range. The 70M configuration also remains close to the default, and the two smaller models exhibit limited degradation. These observations support the use of the 50M configuration as a compact model with performance close to that of the largest tested architecture.

All configurations exhibit decreasing $\mathrm{RT@}k$ and increasing $\mathrm{EM@}k$ as $k$ grows, consistent with the candidate-level and product-level aggregation rules, respectively. Thus, the relationship between forward consistency and reference recovery persists across the tested architectures.

\begin{figure}[!tb]
\centering
\includegraphics[width=0.88\columnwidth]{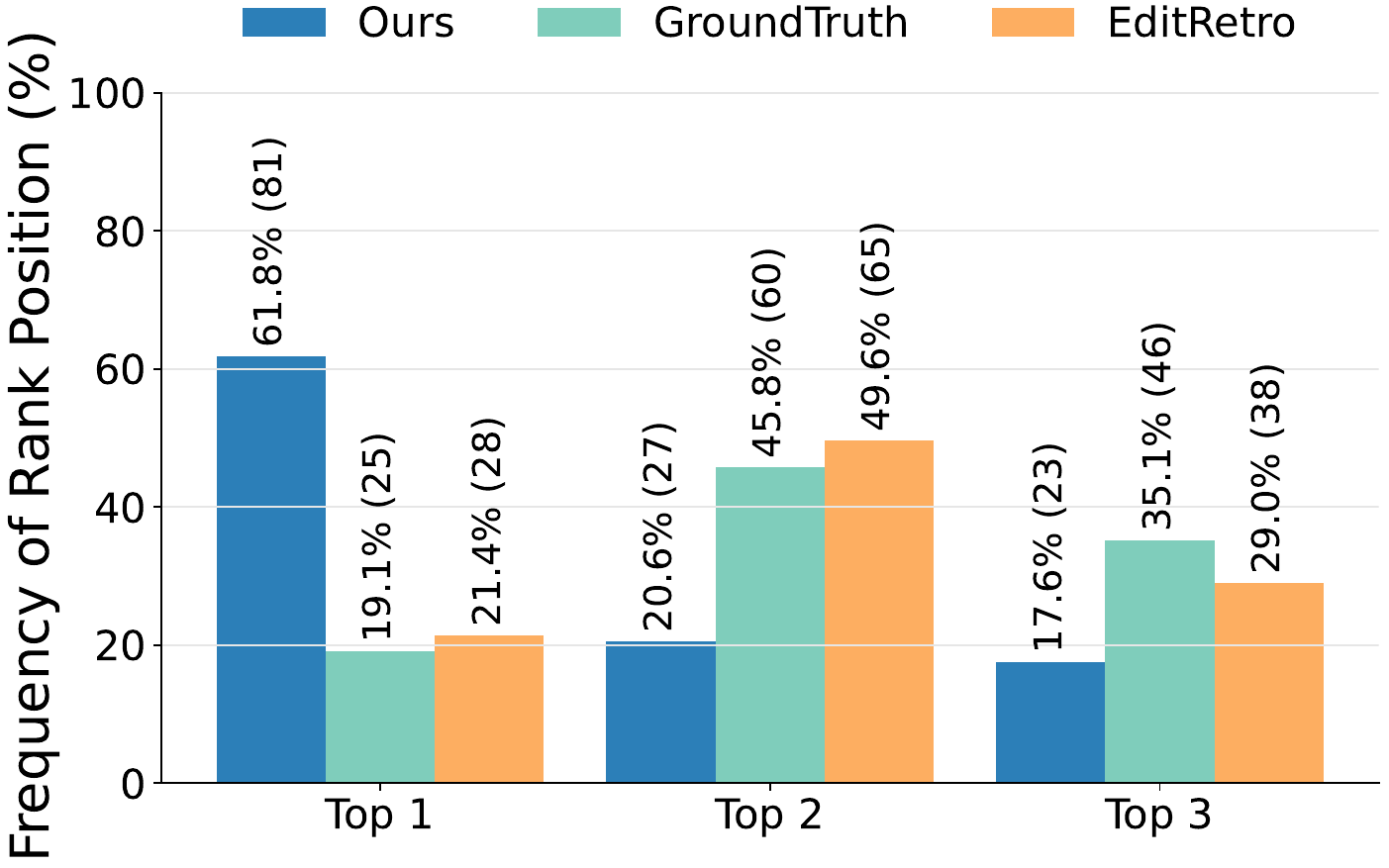}
\caption{Expert-assigned rank positions for RetroRLFF (Ours), recorded reference reactants (GroundTruth), and EditRetro on 131 selected test cases. Top 1, Top 2, and Top 3 denote individual rank positions, not cumulative top-$k$ accuracy. Labels report percentages and case counts; equal ranks are permitted.}
\label{fig:human-rank-distribution}
\end{figure}

\begin{figure}[!tb]
\centering
\includegraphics[width=0.88\columnwidth]{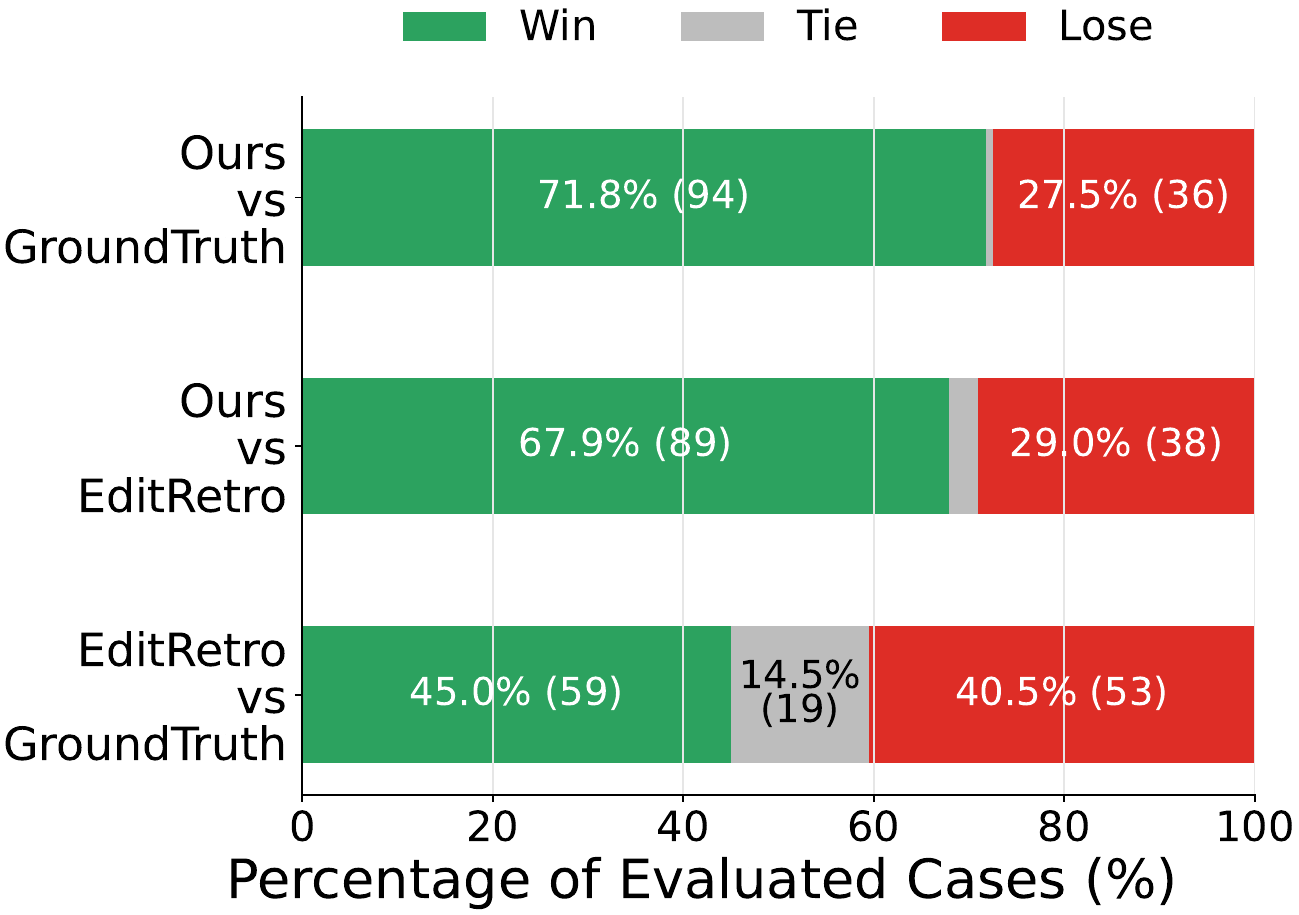}
\caption{Pairwise expert preferences derived from the rankings of 131 test cases. Outcomes are reported from the perspective of the first named reactant set. Green, gray, and red indicate wins, ties, and losses, respectively. Percentages use 131 as the denominator; counts appear in parentheses where labeled.}
\label{fig:human-pairwise-preferences}
\end{figure}

\subsubsection{Human Expert Evaluation}
\label{subsubsec:human-evaluation}


To complement forward-model-based evaluation, we assess predictions on 131 instances randomly selected from the USPTO-50K test set. For each target product, organic synthesis experts compare three reactant sets: the recorded reference, the RetroRLFF prediction, and the EditRetro prediction \cite{TF09}. The sets are anonymized and randomly ordered. Experts rank them by synthetic feasibility, with a better rank indicating a reactant set judged more feasible for synthesizing the target product. Equal ranks are permitted and retained as their assigned numerical positions: sets tied at rank 1 are each counted as Top 1, and sets tied at rank 2 are each counted as Top 2. These positions are reported as individual ranks rather than cumulative top-$k$ accuracy. We summarize the assigned rank positions and derive all three pairwise comparisons from these rankings. A win means that the first set is ranked ahead of the second, a loss means the reverse, and a tie means equal ranks.

Fig.~\ref{fig:human-rank-distribution} shows that RetroRLFF receives first place in 81 cases (61.8\%), compared with 25 (19.1\%) for the reference and 28 (21.4\%) for EditRetro. It is assigned second and third place in 27 (20.6\%) and 23 (17.6\%) cases, respectively. By comparison, the reference and EditRetro are most frequently placed second, in 60 (45.8\%) and 65 (49.6\%) cases. Thus, expert preference on this subset favors RetroRLFF, complementing the forward-consistency results in Table~\ref{tab:benchmark}.

\begin{figure*}[!t]
\centering
\includegraphics[width=\textwidth,height=0.80\textheight,keepaspectratio]{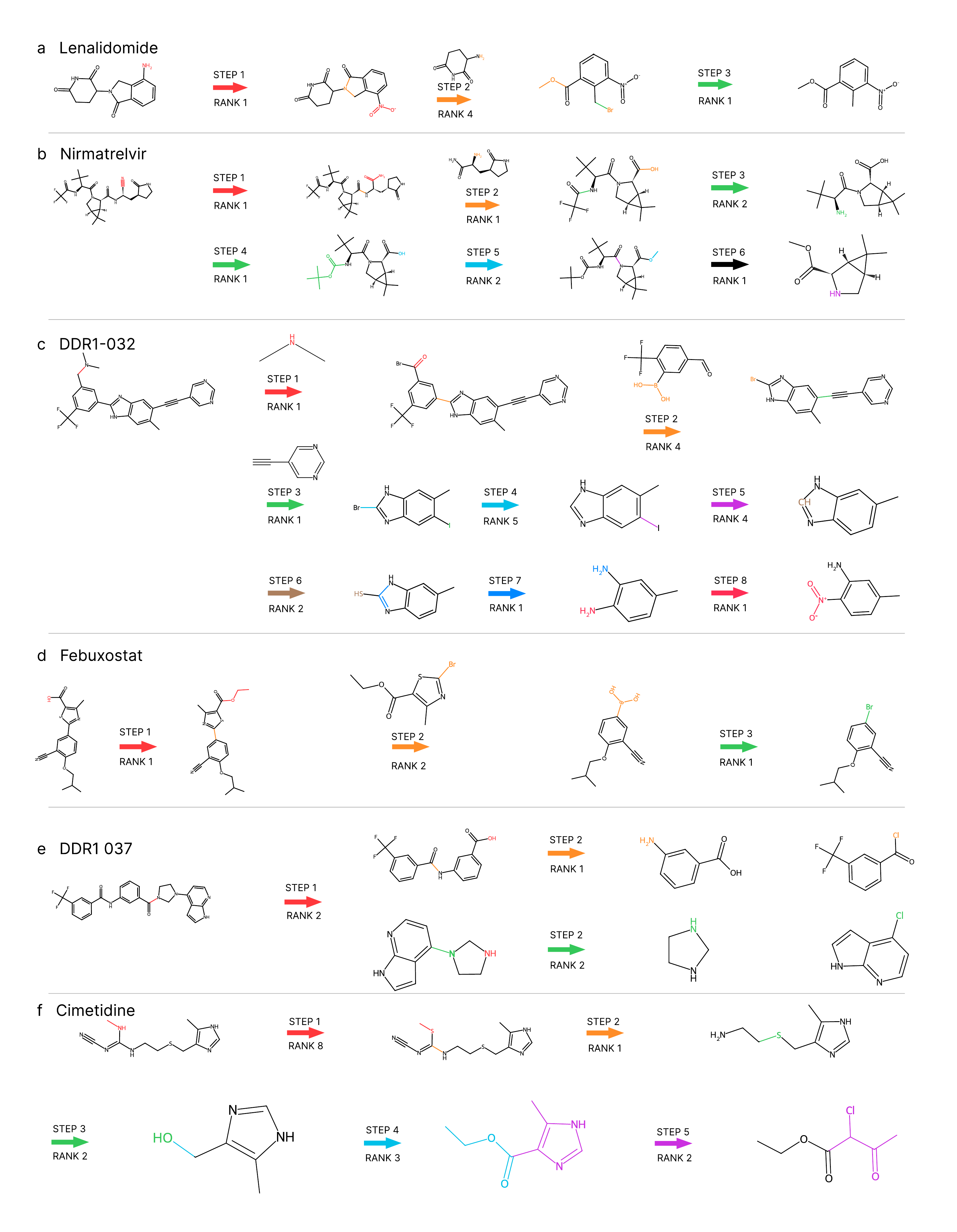}
\caption{Retrosynthetic suggestions for (a) lenalidomide, (b) nirmatrelvir, (c) DDR1-032, (d) febuxostat, (e) DDR1-037, and (f) cimetidine. Arrows point to proposed precursors; STEP and RANK indicate the expansion step and candidate position. Colors highlight structural changes. Panel (e) expands both precursor branches.}
\label{fig:multistep_routes}
\end{figure*}

RetroRLFF versus the reference (Fig.~\ref{fig:human-pairwise-preferences}) yields 94 wins (71.8\%), one tie (0.8\%), and 36 losses (27.5\%). The preference for model predictions over recorded reactants supports the distinction between reference agreement and expert-assessed synthetic feasibility: reproducing the recorded solution is not sufficient to capture relative chemical preference.

RetroRLFF versus EditRetro yields 89 wins (67.9\%), four ties (3.1\%), and 38 losses (29.0\%). Although EditRetro achieves higher exact-match accuracy in Table~\ref{tab:benchmark}, RetroRLFF is preferred more often on the reviewed cases, indicating that reference recovery and expert preference provide complementary assessments.

EditRetro versus the reference yields 59 wins (45.0\%), 19 ties (14.5\%), and 53 losses (40.5\%). This comparatively balanced outcome contrasts with RetroRLFF's larger preference advantage over both alternatives. These findings support the practical relevance of RetroRLFF's predictions within the selected subset.

\subsubsection{Multistep Retrosynthesis Prediction by Our Method}
\label{subsubsec:multistep-retrosynthesis}

We recursively apply RetroRLFF to six target molecules, expanding intermediates through successive single-step predictions. Fig.~\ref{fig:multistep_routes} shows the proposed precursors and their candidate ranks. Transformations are discussed below in the forward synthetic direction.

Lenalidomide (Fig.~\ref{fig:multistep_routes}a) involves three predictions ranked 1, 4, and 1. They suggest nitro-group reduction, lactam construction from an aminoglutarimide and a bromomethyl-substituted nitrobenzoate, and benzylic bromination. The central disconnection requires exploring beyond the first-ranked candidate.

Nirmatrelvir (Fig.~\ref{fig:multistep_routes}b) involves six predictions, all within the top two. The proposed transformations include amide dehydration to the nitrile, amide coupling, trifluoroacetylation, Boc deprotection, and ester hydrolysis. The intermediates retain stereochemical annotations on the bicyclic scaffold, illustrating recursive decomposition involving both coupling and protected precursors.

DDR1-032 (Fig.~\ref{fig:multistep_routes}c) has eight predictions, all within the top five. Later expansions identify alkynyl coupling, halogenated benzimidazole intermediates, heterocycle construction, and nitro-group reduction. However, the acyl bromide depicted after the first expansion and the aldehyde-bearing precursor in the second require clarification before claiming a chemically continuous route.

Febuxostat (Fig.~\ref{fig:multistep_routes}d) involves three predictions ranked 1, 2, and 1. They identify ester hydrolysis, Suzuki coupling between an arylboronic acid and a brominated thiazole ester, and borylation from an aryl bromide. The nitrile and ether substituents remain intact throughout the illustrated sequence.

DDR1-037 (Fig.~\ref{fig:multistep_routes}e) illustrates convergent decomposition. The rank-2 amide disconnection produces a substituted benzoic acid and an azaindole-containing cyclic diamine. Subsequent expansions identify amide-forming precursors at rank 1 and aromatic-substitution precursors at rank 2, exposing both synthetic branches.

Cimetidine (Fig.~\ref{fig:multistep_routes}f) involves five predictions ranked 8, 1, 2, 3, and 2. These suggest cyanoguanidine formation, thioether assembly, ester reduction, and imidazole construction. The rank-8 initial suggestion highlights the potential importance of retaining lower-ranked candidates during recursive exploration.

All 28 illustrated predictions occur within the top 10, including 22 within the top two. These selected cases illustrate diverse precursor relationships, rather than establish a multistep success rate or isolate the effect of RLFF.

\subsubsection{Case Study}
\label{subsubsec:qualitative-routes}

\begin{figure*}[!t]
\centering
\includegraphics[width=\textwidth]{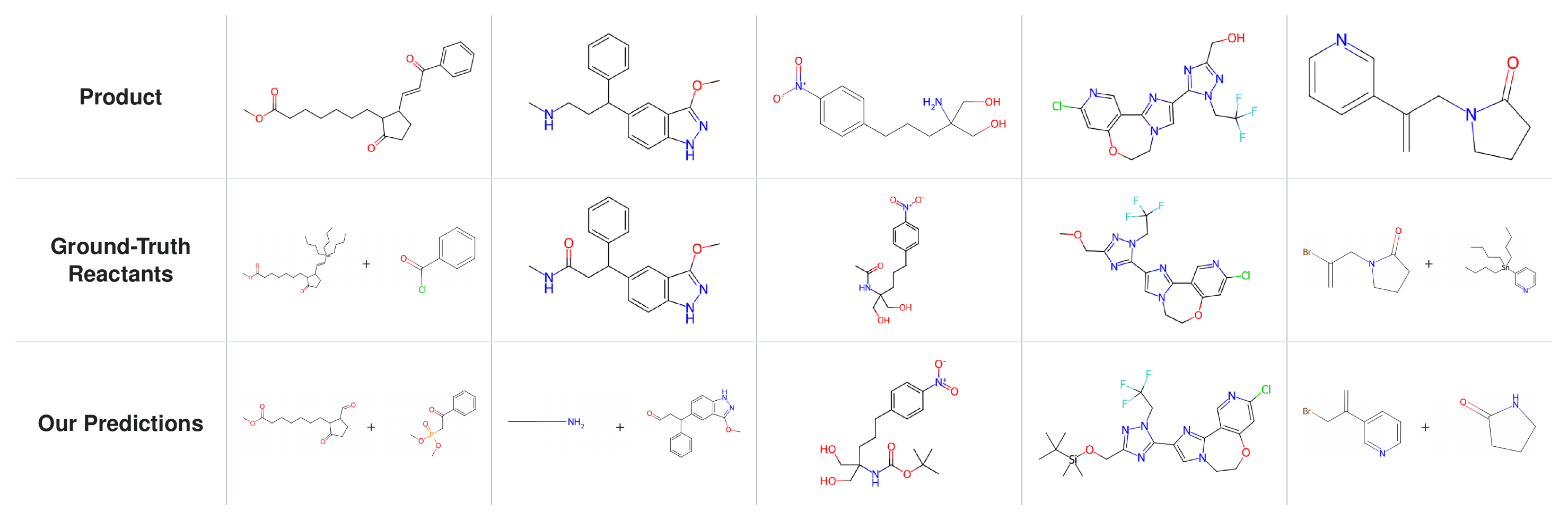}
\caption{Five USPTO-50K test cases. Rows show target products, recorded ground-truth reactants, and RetroRLFF predictions. Columns illustrate enone construction, secondary-amine formation, amino-diol deprotection, heterocyclic alcohol deprotection, and lactam alkylation.}
\label{fig:case_study}
\end{figure*}

Fig.~\ref{fig:case_study} compares five USPTO-50K test products, their recorded reactants, and RetroRLFF predictions. We discuss each column from left to right, focusing on chemically plausible alternatives and potential synthetic advantages.

For the enone-containing product (first column), the reference suggests acylative coupling of a vinylstannane with benzoyl chloride. RetroRLFF proposes Horner--Wadsworth--Emmons olefination of an aldehyde with a $\beta$-ketophosphonate. This alternative avoids an organotin precursor and tin-containing waste, potentially simplifying purification. Chemoselectivity toward the aldehyde over the cyclopentanone and alkene geometry require verification.

For the indazole-containing secondary amine (second column), the reference reduces an $N$-methyl amide, whereas the prediction suggests reductive amination of the corresponding aldehyde with methylamine. The latter offers a recognizable alternative that may permit milder reducing conditions, subject to compatibility with the heterocycle.

For the nitroaryl amino diol (third column), RetroRLFF replaces the reference's $N$-acetyl precursor with a Boc-protected amine. Acid-mediated Boc removal could avoid the more forcing hydrolysis potentially required for deacetylation while retaining the nitro group and both hydroxymethyl groups.

For the heterocyclic alcohol (fourth column), the reference proposes methyl-ether cleavage, whereas RetroRLFF suggests TBS-ether deprotection. Selective silicon--oxygen bond cleavage offers a plausible strategy for exposing the alcohol while preserving the scaffold's other ether linkage, although substrate compatibility remains to be tested.

For the pyridyl-substituted lactam (fifth column), the reference uses Stille coupling with a pyridylstannane. RetroRLFF instead proposes lactam $N$-alkylation with a pyridyl-substituted allylic bromide. This avoids an organotin coupling partner, but competing $O$-alkylation and reactions involving the pyridine nitrogen require assessment.

These cases support evaluating plausible non-reference alternatives beyond exact matching. Without reaction conditions, yields, or precursor-accessibility data, the proposed advantages remain structure-based hypotheses rather than evidence of experimental or overall route superiority.


\section{Conclusion}
\label{sec:conclusion}

RetroRLFF incorporates forward reaction feedback into single-step retrosynthesis to address the mismatch between reference agreement and synthetic feasibility. On USPTO-50K, it improves top-1 round-trip accuracy from 83.1\% to 90.1\% while retaining reference recovery, with expert evaluation supporting the plausibility of its predictions on the reviewed subset.

\subsection{Limitations}
Although the reward balances reference agreement and forward consistency, its feasibility signal remains a surrogate provided by a fixed forward model. Prediction errors and biases can separate round-trip success from experimental outcomes, and expert preference does not replace experimental validation. Moreover, the reward does not explicitly account for precursor cost and availability, catalyst requirements, reaction conditions, or yield. Consequently, forward-consistent predictions are not necessarily economical or practical to execute.

The current policy directly maps product SMILES to reactant SMILES without explicit intermediate chemical analysis or tool use. Its feedback evaluates the generated reactant set, rather than a sequence of chemical decisions. It therefore does not explicitly model how chemists examine bond disconnections, functional-group compatibility, and competing transformations, or consult literature and computational tools before selecting precursors.

\subsection{Future Work}
We will extend the reward toward a weighted, multi-objective assessment combining feasibility with precursor accessibility, material and catalyst costs, reaction conditions, and expected yield. Independent predictors and experimental evidence will be explored to calibrate these signals and reduce dependence on a single forward evaluator. The resulting objective should reflect practical synthesis requirements while making tradeoffs between feasibility and resource use explicit.

A second direction is to extend RetroRLFF into an agent harness that supports iterative analysis and tool use within each single-step prediction. The policy could inspect alternative disconnections, retrieve reaction precedents, query precursor availability, and evaluate candidates through multiple tools, revising its decisions from the returned evidence. Such predictions would form longer decision trajectories, which reinforcement learning could optimize using the comprehensive weighted reward and credit assignment across analysis, tool selection, and final precursor generation.


\bibliographystyle{IEEEtran}
\bibliography{references}

@article{F02,
  author = {Schwaller, Philippe and Petraglia, Riccardo and Zullo, Valerio and Nair, Vishnu H. and Haeuselmann, Rico Andreas and Pisoni, Riccardo and Bekas, Costas and Iuliano, Anna and Laino, Teodoro},
  title = {{Predicting retrosynthetic pathways using transformer-based models and a hyper-graph exploration strategy}},
  journal = {Chemical Science},
  year = {2020},
  volume = {11},
  number = {12},
  pages = {3316--3325},
  doi = {10.1039/c9sc05704h},
  url = {https://doi.org/10.1039/c9sc05704h}
}

@article{T04,
  author = {Chen, Shuan and Jung, Yousung},
  title = {{Deep Retrosynthetic Reaction Prediction using Local Reactivity and Global Attention}},
  journal = {JACS Au},
  year = {2021},
  volume = {1},
  number = {10},
  pages = {1612--1620},
  doi = {10.1021/jacsau.1c00246},
  url = {https://doi.org/10.1021/jacsau.1c00246}
}

@article{T01,
  author = {Coley, Connor W. and Rogers, Luke and Green, William H. and Jensen, Klavs F.},
  title = {{Computer-Assisted Retrosynthesis Based on Molecular Similarity}},
  journal = {ACS Central Science},
  year = {2017},
  volume = {3},
  number = {12},
  pages = {1237--1245},
  doi = {10.1021/acscentsci.7b00355},
  url = {https://doi.org/10.1021/acscentsci.7b00355}
}

@article{T02,
  author = {Segler, Marwin H. S. and Waller, Mark P.},
  title = {{Neural-Symbolic Machine Learning for Retrosynthesis and Reaction Prediction}},
  journal = {Chemistry -- A European Journal},
  year = {2017},
  volume = {23},
  number = {25},
  pages = {5966--5971},
  doi = {10.1002/chem.201605499},
  url = {https://doi.org/10.1002/chem.201605499}
}

@inproceedings{T03,
  author = {Dai, Hanjun and Li, Chengtao and Coley, Connor and Dai, Bo and Song, Le},
  title = {{Retrosynthesis Prediction with Conditional Graph Logic Network}},
  booktitle = {Advances in Neural Information Processing Systems},
  year = {2019},
  volume = {32},
  url = {https://proceedings.neurips.cc/paper_files/paper/2019/file/0d2b2061826a5df3221116a5085a6052-Paper.pdf}
}

@article{T05,
  author = {Seidl, Philipp and Renz, Philipp and Dyubankova, Natalia and Neves, Paulo and Verhoeven, Jonas and Wegner, J\"{o}rg K. and Segler, Marwin and Hochreiter, Sepp and Klambauer, G\"{u}nter},
  title = {{Improving Few- and Zero-Shot Reaction Template Prediction Using Modern Hopfield Networks}},
  journal = {Journal of Chemical Information and Modeling},
  year = {2022},
  volume = {62},
  number = {9},
  pages = {2111--2120},
  doi = {10.1021/acs.jcim.1c01065},
  url = {https://doi.org/10.1021/acs.jcim.1c01065}
}

@article{T06,
  author = {Xie, Shufang and Yan, Rui and Guo, Junliang and Xia, Yingce and Wu, Lijun and Qin, Tao},
  title = {{Retrosynthesis Prediction with Local Template Retrieval}},
  journal = {Proceedings of the AAAI Conference on Artificial Intelligence},
  year = {2023},
  volume = {37},
  number = {4},
  pages = {5330--5338},
  doi = {10.1609/aaai.v37i4.25664},
  url = {https://doi.org/10.1609/aaai.v37i4.25664}
}

@inproceedings{ST01,
  author = {Shi, Chence and Xu, Minkai and Guo, Hongyu and Zhang, Ming and Tang, Jian},
  title = {{A Graph to Graphs Framework for Retrosynthesis Prediction}},
  booktitle = {Proceedings of the 37th International Conference on Machine Learning},
  year = {2020},
  volume = {119},
  series = {Proceedings of Machine Learning Research},
  pages = {8818--8827},
  publisher = {PMLR},
  url = {https://proceedings.mlr.press/v119/shi20d.html}
}

@inproceedings{ST02,
  author = {Yan, Chaochao and Ding, Qianggang and Zhao, Peilin and Zheng, Shuangjia and Yang, Jinyu and Yu, Yang and Huang, Junzhou},
  title = {{RetroXpert: Decompose Retrosynthesis Prediction Like A Chemist}},
  booktitle = {Advances in Neural Information Processing Systems},
  year = {2020},
  volume = {33},
  pages = {11248--11258},
  url = {https://proceedings.neurips.cc/paper_files/paper/2020/file/819f46e52c25763a55cc642422644317-Paper.pdf}
}

@inproceedings{ST03,
  author = {Somnath, Vignesh Ram and Bunne, Charlotte and Coley, Connor and Krause, Andreas and Barzilay, Regina},
  title = {{Learning Graph Models for Retrosynthesis Prediction}},
  booktitle = {Advances in Neural Information Processing Systems},
  year = {2021},
  volume = {34},
  pages = {9405--9415},
  url = {https://proceedings.neurips.cc/paper_files/paper/2021/file/4e2a6330465c8ffcaa696a5a16639176-Paper.pdf}
}

@article{TF01,
  author = {Liu, Bowen and Ramsundar, Bharath and Kawthekar, Prasad and Shi, Jade and Gomes, Joseph and Luu Nguyen, Quang and Ho, Stephen and Sloane, Jack and Wender, Paul and Pande, Vijay},
  title = {{Retrosynthetic Reaction Prediction Using Neural Sequence-to-Sequence Models}},
  journal = {ACS Central Science},
  year = {2017},
  volume = {3},
  number = {10},
  pages = {1103--1113},
  doi = {10.1021/acscentsci.7b00303},
  url = {https://doi.org/10.1021/acscentsci.7b00303}
}

@article{TF02,
  author = {Zheng, Shuangjia and Rao, Jiahua and Zhang, Zhongyue and Xu, Jun and Yang, Yuedong},
  title = {{Predicting Retrosynthetic Reactions Using Self-Corrected Transformer Neural Networks}},
  journal = {Journal of Chemical Information and Modeling},
  year = {2020},
  volume = {60},
  number = {1},
  pages = {47--55},
  doi = {10.1021/acs.jcim.9b00949},
  url = {https://doi.org/10.1021/acs.jcim.9b00949}
}

@article{TF03,
  author = {Sacha, Miko{\l}aj and B{\l}a\.{z}, Miko{\l}aj and Byrski, Piotr and D\k{a}browski-Tuma\'{n}ski, Pawe{\l} and Chromi\'{n}ski, Miko{\l}aj and Loska, Rafa{\l} and W{\l}odarczyk-Pruszy\'{n}ski, Pawe{\l} and Jastrz\k{e}bski, Stanis{\l}aw},
  title = {{Molecule Edit Graph Attention Network: Modeling Chemical Reactions as Sequences of Graph Edits}},
  journal = {Journal of Chemical Information and Modeling},
  year = {2021},
  volume = {61},
  number = {7},
  pages = {3273--3284},
  doi = {10.1021/acs.jcim.1c00537},
  url = {https://doi.org/10.1021/acs.jcim.1c00537}
}

@article{TF04,
  author = {Tu, Zhengkai and Coley, Connor W.},
  title = {{Permutation Invariant Graph-to-Sequence Model for Template-Free Retrosynthesis and Reaction Prediction}},
  journal = {Journal of Chemical Information and Modeling},
  year = {2022},
  volume = {62},
  number = {15},
  pages = {3503--3513},
  doi = {10.1021/acs.jcim.2c00321},
  url = {https://doi.org/10.1021/acs.jcim.2c00321}
}

@article{TF05,
  author = {Zhong, Zipeng and Song, Jie and Feng, Zunlei and Liu, Tiantao and Jia, Lingxiang and Yao, Shaolun and Wu, Min and Hou, Tingjun and Song, Mingli},
  title = {{Root-aligned SMILES: a tight representation for chemical reaction prediction}},
  journal = {Chemical Science},
  year = {2022},
  volume = {13},
  number = {31},
  pages = {9023--9034},
  doi = {10.1039/d2sc02763a},
  url = {https://doi.org/10.1039/d2sc02763a}
}

@inproceedings{TF06,
  author = {Wan, Yue and Hsieh, Chang-Yu and Liao, Ben and Zhang, Shengyu},
  title = {{Retroformer: Pushing the Limits of End-to-end Retrosynthesis Transformer}},
  booktitle = {Proceedings of the 39th International Conference on Machine Learning},
  year = {2022},
  volume = {162},
  series = {Proceedings of Machine Learning Research},
  pages = {22475--22490},
  publisher = {PMLR},
  url = {https://proceedings.mlr.press/v162/wan22a.html}
}

@article{N09,
  author = {Irwin, Ross and Dimitriadis, Spyridon and He, Jiazhen and Bjerrum, Esben Jannik},
  title = {{Chemformer: a pre-trained transformer for computational chemistry}},
  journal = {Machine Learning: Science and Technology},
  year = {2022},
  volume = {3},
  number = {1},
  note = {Article 015022},
  doi = {10.1088/2632-2153/ac3ffb},
  url = {https://doi.org/10.1088/2632-2153/ac3ffb}
}

@inproceedings{N04,
  author = {Igashov, Ilia and Schneuing, Arne and Segler, Marwin and Bronstein, Michael and Correia, Bruno},
  title = {{RetroBridge: Modeling Retrosynthesis with Markov Bridges}},
  booktitle = {Proceedings of the Twelfth International Conference on Learning Representations},
  year = {2024},
  pages = {39622--39640},
  url = {https://proceedings.iclr.cc/paper_files/paper/2024/file/acb94e709f02895fd98b5867f0b184f3-Paper-Conference.pdf}
}

@article{N05,
  author = {Current, Sean and Chen, Ziqi and Adu-Ampratwum, Daniel and Ning, Xia and Parthasarathy, Srinivasan},
  title = {{DiffER: categorical diffusion ensembles for single-step chemical retrosynthesis}},
  journal = {Journal of Cheminformatics},
  year = {2025},
  volume = {17},
  number = {1},
  note = {Article 112},
  doi = {10.1186/s13321-025-01056-7},
  url = {https://doi.org/10.1186/s13321-025-01056-7}
}

@inproceedings{TF11,
  author = {Yadav, Robin and Yan, Qi and Wolf, Guy and Bose, Joey and Liao, Renjie},
  title = {{RETRO SYNFLOW: Discrete Flow-Matching for Accurate and Diverse Single-Step Retrosynthesis}},
  booktitle = {Advances in Neural Information Processing Systems},
  year = {2025},
  volume = {38},
  pages = {55229--55258},
  doi = {10.52202/085713-1846},
  url = {https://proceedings.neurips.cc/paper_files/paper/2025/file/4f86833d5cc98ec32e470ef1c8cb82e3-Paper-Conference.pdf}
}

@article{ST04,
  author = {Wang, Xiaorui and Li, Yuquan and Qiu, Jiezhong and Chen, Guangyong and Liu, Huanxiang and Liao, Benben and Hsieh, Chang-Yu and Yao, Xiaojun},
  title = {{RetroPrime: A Diverse, plausible and Transformer-based method for Single-Step retrosynthesis predictions}},
  journal = {Chemical Engineering Journal},
  year = {2021},
  volume = {420},
  note = {Article 129845},
  doi = {10.1016/j.cej.2021.129845},
  url = {https://doi.org/10.1016/j.cej.2021.129845}
}

@article{ST05,
  author = {Chen, Ziqi and Ayinde, Oluwatosin R. and Fuchs, James R. and Sun, Huan and Ning, Xia},
  title = {{G2Retro as a two-step graph generative models for retrosynthesis prediction}},
  journal = {Communications Chemistry},
  year = {2023},
  volume = {6},
  number = {1},
  note = {Article 102},
  doi = {10.1038/s42004-023-00897-3},
  url = {https://doi.org/10.1038/s42004-023-00897-3}
}

@inproceedings{RL02,
  author = {Ouyang, Long and Wu, Jeffrey and Jiang, Xu and Almeida, Diogo and Wainwright, Carroll and Mishkin, Pamela and Zhang, Chong and Agarwal, Sandhini and Slama, Katarina and Ray, Alex and Schulman, John and Hilton, Jacob and Kelton, Fraser and Miller, Luke and Simens, Maddie and Askell, Amanda and Welinder, Peter and Christiano, Paul F. and Leike, Jan and Lowe, Ryan},
  title = {{Training language models to follow instructions with human feedback}},
  booktitle = {Advances in Neural Information Processing Systems},
  year = {2022},
  volume = {35},
  pages = {27730--27744},
  doi = {10.52202/068431-2011},
  url = {https://proceedings.neurips.cc/paper_files/paper/2022/file/b1efde53be364a73914f58805a001731-Paper-Conference.pdf}
}

@article{F01,
  author = {Schwaller, Philippe and Laino, Teodoro and Gaudin, Th\'{e}ophile and Bolgar, Peter and Hunter, Christopher A. and Bekas, Costas and Lee, Alpha A.},
  title = {{Molecular Transformer: A Model for Uncertainty-Calibrated Chemical Reaction Prediction}},
  journal = {ACS Central Science},
  year = {2019},
  volume = {5},
  number = {9},
  pages = {1572--1583},
  doi = {10.1021/acscentsci.9b00576},
  url = {https://doi.org/10.1021/acscentsci.9b00576}
}

@article{ST06,
  author = {Zhong, Weihe and Yang, Ziduo and Chen, Calvin Yu-Chian},
  title = {{Retrosynthesis prediction using an end-to-end graph generative architecture for molecular graph editing}},
  journal = {Nature Communications},
  year = {2023},
  volume = {14},
  number = {1},
  note = {Article 3009},
  doi = {10.1038/s41467-023-38851-5},
  url = {https://doi.org/10.1038/s41467-023-38851-5}
}

@inproceedings{DAPO,
  author = {Liu, Jiacai and Wang, Chaojie and Liu, Chris and Zeng, Liang and Yan, Rui and Sun, Yiwen and Liu, Yang},
  title = {{DAPO : Improving Multi-Step Reasoning Abilities of Large Language Models with Direct Advantage-Based Policy Optimization}},
  booktitle = {Advances in Neural Information Processing Systems},
  year = {2025},
  volume = {38},
  pages = {71445--71477},
  doi = {10.52202/085713-2401},
  url = {https://proceedings.neurips.cc/paper_files/paper/2025/file/6789f033ebde742552e5db84fb5d414a-Paper-Conference.pdf}
}

@article{ST09,
  author = {Zhao, Peng-Cheng and Wei, Xue-Xin and Wang, Qiong and Wang, Qi-Hao and Li, Jia-Ning and Shang, Jie and Lu, Cheng and Shi, Jian-Yu},
  title = {{Single-step retrosynthesis prediction via multitask graph representation learning}},
  journal = {Nature Communications},
  year = {2025},
  volume = {16},
  number = {1},
  note = {Article 814},
  doi = {10.1038/s41467-025-56062-y},
  url = {https://doi.org/10.1038/s41467-025-56062-y}
}

@article{TF10,
  author = {Tetko, Igor V. and Karpov, Pavel and Van Deursen, Ruud and Godin, Guillaume},
  title = {{State-of-the-art augmented NLP transformer models for direct and single-step retrosynthesis}},
  journal = {Nature Communications},
  year = {2020},
  volume = {11},
  number = {1},
  note = {Article 5575},
  doi = {10.1038/s41467-020-19266-y},
  url = {https://doi.org/10.1038/s41467-020-19266-y}
}

@article{TF08,
  author = {Lin, Zaiyun and Yin, Shiqiu and Shi, Lei and Zhou, Wenbiao and Zhang, Yingsheng John},
  title = {{G2GT: Retrosynthesis Prediction with Graph-to-Graph Attention Neural Network and Self-Training}},
  journal = {Journal of Chemical Information
and Modeling},
  year = {2023},
  volume = {63},
  number = {7},
  pages = {1894--1905},
  doi = {10.1021/acs.jcim.2c01302},
  url = {https://doi.org/10.1021/acs.jcim.2c01302}
}

@article{TF09,
  author = {Han, Yuqiang and Xu, Xiaoyang and Hsieh, Chang-Yu and Ding, Keyan and Xu, Hongxia and Xu, Renjun and Hou, Tingjun and Zhang, Qiang and Chen, Huajun},
  title = {{Retrosynthesis prediction with an iterative string editing model}},
  journal = {Nature Communications},
  year = {2024},
  volume = {15},
  number = {1},
  note = {Article 6404},
  doi = {10.1038/s41467-024-50617-1},
  url = {https://doi.org/10.1038/s41467-024-50617-1}
}

@inproceedings{RL07,
  author = {Christiano, Paul F. and Leike, Jan and Brown, Tom and Martic, Miljan and Legg, Shane and Amodei, Dario},
  title = {{Deep Reinforcement Learning from Human Preferences}},
  booktitle = {Advances in Neural Information Processing Systems},
  year = {2017},
  volume = {30},
  url = {https://proceedings.neurips.cc/paper_files/paper/2017/file/d5e2c0adad503c91f91df240d0cd4e49-Paper.pdf}
}

@inproceedings{RL08,
  author = {Stiennon, Nisan and Ouyang, Long and Wu, Jeffrey and Ziegler, Daniel and Lowe, Ryan and Voss, Chelsea and Radford, Alec and Amodei, Dario and Christiano, Paul F.},
  title = {{Learning to summarize with human feedback}},
  booktitle = {Advances in Neural Information Processing Systems},
  year = {2020},
  volume = {33},
  pages = {3008--3021},
  url = {https://proceedings.neurips.cc/paper_files/paper/2020/file/1f89885d556929e98d3ef9b86448f951-Paper.pdf}
}

@misc{RL01,
  author = {Schulman, John and Wolski, Filip and Dhariwal, Prafulla and Radford, Alec and Klimov, Oleg},
  title = {{Proximal Policy Optimization Algorithms}},
  year = {2017},
  eprint = {1707.06347},
  archivePrefix = {arXiv},
  primaryClass = {cs.LG},
  doi = {10.48550/arXiv.1707.06347},
  url = {https://arxiv.org/abs/1707.06347},
  note = {arXiv preprint}
}

@inproceedings{RL06,
  author = {Lee, Harrison and Phatale, Samrat and Mansoor, Hassan and Mesnard, Thomas and Ferret, Johan and Lu, Kellie Ren and Bishop, Colton and Hall, Ethan and Carbune, Victor and Rastogi, Abhinav and Prakash, Sushant},
  title = {{{RLAIF} vs. {RLHF}: Scaling Reinforcement Learning from Human Feedback with {AI} Feedback}},
  booktitle = {Proceedings of the 41st International Conference on Machine Learning},
  year = {2024},
  volume = {235},
  series = {Proceedings of Machine Learning Research},
  pages = {26874--26901},
  publisher = {PMLR},
  url = {https://proceedings.mlr.press/v235/lee24t.html}
}

@inproceedings{RL09,
  author = {Gao, Leo and Schulman, John and Hilton, Jacob},
  title = {{Scaling Laws for Reward Model Overoptimization}},
  booktitle = {Proceedings of the 40th International Conference on Machine Learning},
  year = {2023},
  volume = {202},
  series = {Proceedings of Machine Learning Research},
  pages = {10835--10866},
  publisher = {PMLR},
  url = {https://proceedings.mlr.press/v202/gao23h.html}
}

@article{RL03,
  author = {Baker, Frazier N. and Chen, Ziqi and Adu-Ampratwum, Daniel and Ning, Xia},
  title = {{RLSynC: Offline--Online Reinforcement Learning for Synthon Completion}},
  journal = {Journal of Chemical Information and Modeling},
  year = {2024},
  volume = {64},
  number = {17},
  pages = {6723--6735},
  doi = {10.1021/acs.jcim.4c00554},
  url = {https://doi.org/10.1021/acs.jcim.4c00554}
}

@article{N03,
  author = {Gai\'{n}ski, Piotr and Koziarski, Micha{\l} and Maziarz, Krzysztof and Segler, Marwin and Tabor, Jacek and \'{S}mieja, Marek},
  title = {{Diverse and feasible retrosynthesis using GFlowNets}},
  journal = {Information Sciences},
  year = {2025},
  volume = {714},
  note = {Article 122194},
  doi = {10.1016/j.ins.2025.122194},
  url = {https://doi.org/10.1016/j.ins.2025.122194}
}

@inproceedings{RL04,
  author = {Narayanan, Siddharth and Braza, James and Griffiths, Ryan-Rhys and Bou, Albert and Wellawatte, Geemi and Caldas Ramos, Mayk and Mitchener, Ludovico and Pieler, Michael and Rodriques, Sam and White, Andrew},
  title = {{Training a Scientific Reasoning Model for Chemistry}},
  booktitle = {Advances in Neural Information Processing Systems},
  year = {2025},
  volume = {38},
  pages = {157671--157710},
  doi = {10.52202/085713-5269},
  url = {https://proceedings.neurips.cc/paper_files/paper/2025/file/e754c82afb9bb0150bd6b71f5f49a9e6-Paper-Conference.pdf}
}

@misc{RL05,
  author = {Kong, Lecheng and Wang, Xiyuan and Chen, Yixin and Zhang, Muhan},
  title = {{Round-trip Reinforcement Learning: Self-Consistent Training for Better Chemical LLMs}},
  year = {2025},
  eprint = {2510.01527},
  archivePrefix = {arXiv},
  primaryClass = {cs.LG},
  doi = {10.48550/arXiv.2510.01527},
  url = {https://arxiv.org/abs/2510.01527},
  note = {arXiv preprint}
}

@article{GTA,
  author = {Seo, Seung-Woo and Song, You Young and Yang, June Yong and Bae, Seohui and Lee, Hankook and Shin, Jinwoo and Hwang, Sung Ju and Yang, Eunho},
  title = {{GTA: Graph Truncated Attention for Retrosynthesis}},
  journal = {Proceedings of the AAAI Conference on Artificial Intelligence},
  year = {2021},
  volume = {35},
  number = {1},
  pages = {531--539},
  doi = {10.1609/aaai.v35i1.16131},
  url = {https://doi.org/10.1609/aaai.v35i1.16131}
}

@article{lan2025retrosynthesis,
  title={Retrosynthesis prediction via search in (hyper) graph},
  author={Lan, Zixun and Hong, Binjie and Xu, Maochun and Zeng, Zuo and Liu, Zhenfu and Yu, Limin and Ma, Fei},
  journal={IEEE Transactions on Automation Science and Engineering},
  volume={22},
  pages={17749--17760},
  year={2025},
  publisher={IEEE}
}

@article{lan2024rcsearcher,
  title={Rcsearcher: Reaction center identification in retrosynthesis via deep q-learning},
  author={Lan, Zixun and Zeng, Zuo and Hong, Binjie and Liu, Zhenfu and Ma, Fei},
  journal={Pattern Recognition},
  volume={150},
  pages={110318},
  year={2024},
  publisher={Elsevier}
}

@inproceedings{zhu2023single,
  title={Single-step retrosynthesis via reaction center and leaving groups prediction},
  author={Zhu, Jiajun and Hong, Binjie and Lan, Zixun and Ma, Fei},
  booktitle={2023 16th International Congress on Image and Signal Processing, BioMedical Engineering and Informatics (CISP-BMEI)},
  pages={1--5},
  year={2023},
  organization={IEEE}
}

\end{document}